\documentclass[10pt,twocolumn,letterpaper]{article}

\usepackage[pagenumbers]{wacv} 

\usepackage{graphicx}
\usepackage{booktabs}

\usepackage{url}

\usepackage[utf8]{inputenc} 
\usepackage[T1]{fontenc}    
\usepackage{url}            
\usepackage{booktabs}       
\usepackage{amsfonts}       
\usepackage{nicefrac}       
\usepackage{microtype}      
\usepackage{xcolor}         
\usepackage{tabularx}
\usepackage{multirow}
\usepackage{rotating}
\usepackage{amssymb}
\usepackage{amsmath}
\usepackage{multirow}
\usepackage{arydshln}
\usepackage{svg}
\usepackage{listings}
\usepackage{xcolor}
\usepackage{bbm}
\usepackage{tcolorbox}
\usepackage{setspace}
\usepackage{caption}
\usepackage{colortbl,xcolor}

\lstdefinelanguage{JSON}{
    string=[s]{"}{"},
    stringstyle=\color{red},
    comment=[l]{:},
    commentstyle=\color{blue},
    numbers=left,
    numberstyle=\tiny,
    stepnumber=1,
    numbersep=8pt,
    showstringspaces=false,
    breaklines=true,
    frame=single,
    basicstyle=\ttfamily\small,
    keywordstyle=\color{blue},
    identifierstyle=\color{black},
}

\DeclareMathOperator*{\argmin}{arg\,min}
\usepackage{rotating}
\usepackage{amssymb}
\usepackage{multirow}
\usepackage{arydshln}
\usepackage{svg}
\usepackage{listings}
\usepackage{xcolor}
\usepackage{bbm}
\usepackage{tcolorbox}
\usepackage{setspace}

\usepackage{pifont}
\usepackage{graphicx}
\usepackage{multirow} 
\usepackage{amsmath}
\usepackage{adjustbox}
\usepackage{arydshln}
\usepackage{float}
\usepackage{amssymb}
\usepackage{xspace}
\usepackage{makecell}
\usepackage{wrapfig}
\usepackage{comment}
\usepackage{xcolor}
\usepackage{wrapfig}

\usepackage{algorithm,algpseudocode}

\definecolor{wacvblue}{rgb}{0.21,0.49,0.74}
\usepackage[pagebackref,breaklinks,colorlinks,allcolors=wacvblue]{hyperref}

\def\wacvPaperID{571} 
\def\confName{WACV}
\def\confYear{2027}

\title{ReSpace: Text-Driven Autoregressive 3D Indoor Scene Synthesis and Editing}

\author{%
  \textbf{Martin JJ. Bucher} \\
  Stanford University \\
  \and
  \textbf{Iro Armeni} \\
  Stanford University \\
}

\begin{document}


\twocolumn[{%
\renewcommand\twocolumn[1][]{#1}%
\maketitle

\vspace{-12mm}

{\centering
\href{https://respace.mnbucher.com}{\texttt{respace.mnbucher.com}}\par}
\vspace{-2mm}

\begin{center}
    \includegraphics[width=0.80\textwidth]{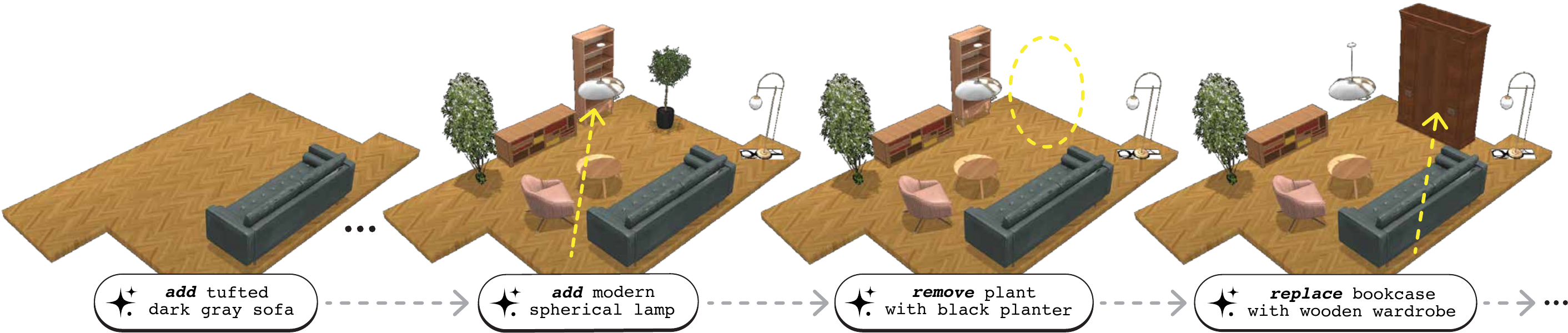}
    \vspace{-1mm}
    \captionof{figure}{We introduce a text-driven framework for 3D indoor scene synthesis and editing—supporting addition, removal, and swapping.
    }
    \label{fig:teaser}
\end{center}
}]

\vspace*{-5mm}

\begin{abstract}
Scene synthesis and editing has emerged as a promising direction in computer graphics. Current trained approaches for 3D indoor scene generation either oversimplify object semantics through one-hot class encodings (e.g., `chair' or `table'), require masked diffusion for editing, ignore room boundaries, or rely on floor plan renderings that fail to capture complex layouts. LLM-based methods enable richer semantics via natural language, but lack editing functionality, are limited to rectangular layouts, or rely on weak spatial reasoning from implicit world models. We introduce \textsc{ReSpace}, a generative framework for autoregressive text-driven 3D indoor scene synthesis and editing. Our approach features a compact structured scene representation with explicit room boundaries that enables asset-agnostic deployment and frames scene manipulation as a next-token prediction task, supporting object addition, removal, and swapping via natural language. We employ supervised fine-tuning with a preference alignment stage to train a specialized language model for object addition that accounts for user instructions, spatial geometry, object semantics, and scene-level composition. We further introduce a voxelization-based evaluation metric capturing fine-grained geometric violations beyond 3D bounding boxes. Experiments surpass state-of-the-art on object addition and achieve superior human-perceived quality on the application of full scene synthesis, despite not being trained on it.
\end{abstract}

\vspace{-4mm}

\section{Introduction}
\label{chap:intro}
Scene synthesis for 3D environments has been a long standing challenge in computer graphics for many decades. In particular, indoor scenes have been of interest due to the wide range of applications in virtual and mixed reality, robotics, entertainment, retail, virtual staging, interior design, and more. As the manual creation and editing of such scenes requires substantial human effort and expertise, significant effort has been put on automating this process. With early approaches revolving around heuristics-based methods and procedural modeling \cite{qi2018human, weiss2018fast, xu2002constraint, yu2011make, purkait2020sg, fisher2015activity}, recent effort has shifted towards deep generative models, i.e., learning the distribution of indoor scenes directly from data. For instance, autoregressive models \cite{paschalidou2021atiss, wang2018deep, ritchie2019fast} learn to stochastically predict a sequence of objects for full scene generation or completion. Another line of work explores diffusion-based models for scene synthesis \cite{tang2024diffuscene, hu2024mixed, maillard2024debara} by learning how to gradually denoise object properties from random noise. Methods based on scene graphs have also been proposed, either by assuming a high-level scene graph as input or by generating a scene graph in a first stage and then obtaining object properties via diffusion \cite{zhai2024echoscene, lin2024instructscene}. With the recent advent of instruction-tuned Large Language Models (LLMs), agent-based approaches have been pursued, relying primarily on the inherent world model of LLMs \cite{sun2024layoutvlm, ccelen2024design, yang2024holodeck, feng2023layoutgpt}. However, existing methods face several limitations: they simplify object semantics via one-hot class labels, ignore room boundaries or rely on fixed-resolution floor plans regardless of scene complexity, depend on zero-shot LLMs or external optimization algorithms given nascent spatial foundation models, and lack direct natural language modification capabilities. While full scene synthesis from scratch is a useful benchmark, practical interior design is inherently iterative --- users refine and adjust scenes progressively, making fine-grained text-driven editing a more valuable capability than one-shot generation. Yet, existing methods for explicit 3D scene representations lack fine-grained text-driven editing with semantic modularity, where individual objects can be added, removed, or swapped independently while preserving scene context.

We propose \textsc{ReSpace}, a novel framework leveraging natural language for intuitive scene synthesis and editing through text commands for object addition, removal, and swapping. We frame scene manipulation as a next-token prediction task, enabled by a structured scene representation (SSR) in a JSON format that represents spatial information (i.e., room boundaries, object semantics, and precise placement) explicitly, and train a specialized model, SG-LLM, for object addition. SG-LLM takes a short object prompt as input and performs single object placement as output. It is prompted by a zero-shot LLM that serves as the user interface, decomposing user instructions into atomic additions/removals. For removals, the zero-shot LLM handles these directly through SSR text editing. We decouple 3D asset selection from the SSR using a stochastic sampler that matches both size and semantics from an existing catalog. Example prompt-output pairs are shown in Fig. \ref{fig:overview} alongside a summary. For evaluation, we introduce a voxelization-based metric capturing fine-grained geometric interactions beyond bounding boxes, quantifying realistic placements such as chairs partially under tables (Fig. \ref{fig:lay-viol} (C)). After Supervised Fine-Tuning (SFT) on instructions, we explore this metric—alongside other constraints—as verifiable reward for preference alignment on SG-LLM. As an application, and to compare with methods that only perform full scene synthesis, SG-LLM processes autoregressively object lists generated by the zero-shot LLM. Experiments on the 3D-FRONT \cite{fu20213d_front} dataset demonstrate a new state-of-the-art for object placement and superior human-perceived quality for full scene synthesis. Code and dataset are available \href{https://github.com/gradientspaces/respace}{here}. In summary, our contributions are as follows:



\begin{itemize}
    \item We present \textsc{ReSpace}, a novel method for controllable indoor scene synthesis and editing, framing object addition and removal via next-token prediction.
    \item We present a supervised fine-tuning pipeline for object addition that surpasses state-of-the-art on placement and achieves strong results for full scene synthesis, with exploratory preference-alignment experiments using Reinforcement Learning with Verifiable Rewards (RLVR).
    \item We introduce a lightweight and interpretable structured scene representation with natural-language object descriptions and explicit numerical values for scene boundaries and object positioning, enabling direct editing and asset-agnostic deployment across 3D catalogs.
    \item We propose the Voxelization-Based Loss (VBL), a novel \textit{evaluation} metric capturing fine-grained geometric interactions beyond 3D bounding boxes (e.g., chair/table).
\end{itemize}

\section{Related Work}
\label{chap:relwork}

\paragraph{3D Indoor Scene Synthesis.} Early approaches relied on heuristics and procedural modeling \cite{qi2018human, weiss2018fast, xu2002constraint, yu2011make, purkait2020sg, fisher2015activity}. With deep learning's emergence, transformers \cite{ritchie2019fast, wang2021sceneformer, paschalidou2021atiss} and diffusion models \cite{tang2024diffuscene, hu2024mixed, wei2023lego, maillard2024debara} gained prominence. Deep Priors \cite{wang2018deep} introduced CNN-based attribute prediction, while Fast\&Flexible \cite{ritchie2019fast} developed a chained CNN pipeline conditionable on floor plan images. SceneFormer \cite{wang2021sceneformer} proposed autoregressive transformers conditioned on floor plans and text descriptions, while ATISS \cite{paschalidou2021atiss} pioneered treating scenes as unordered object sets. FOREST2SEQ \cite{sun2024forest2seq} explores ordering strategies for autoregressive synthesis to improve placement quality. Recent advances include diffusion-based approaches like DiffuScene \cite{tang2024diffuscene}, Mi-Diff \cite{hu2024mixed} (supporting floor plan conditioning via PointNet features), PhyScene \cite{yang2024physcene} (focusing on physically interactable synthesis), and LEGO-Net \cite{wei2023lego} (via rearrangement). Alternative approaches generate unified scene representations: Text2Room \cite{hollein2023text2room} extracts textured meshes from 2D models, DreamScene \cite{li2024dreamscene} uses Gaussian-based text-to-3D generation, and Set-the-Scene \cite{cohen2023set} enables controllable NeRF scenes. Human-centric approaches include MIME \cite{yi2023mime} and SUMMON \cite{ye2022scene}, while scene graph methods \cite{dhamo2021graph, lin2024instructscene, zhai2023commonscenes, zhai2024echoscene, luo2020end} like InstructScene \cite{lin2024instructscene} and EchoScene \cite{zhai2024echoscene} use intermediate graph representations, with EditRoom \cite{zheng2025editroom} extending graph diffusion to language-guided layout editing. Despite these advances, most methods either generate unified representations limiting asset flexibility, focus on end-to-end synthesis without granular editing capabilities, or lack explicit 3D boundary handling for complex layouts and intuitive text-driven manipulation.

\begin{table}
\centering
\caption{Comparison of key properties across recent methods.}
\vspace{-2mm}
\label{tab:method_comparison}
\scriptsize
\setlength{\tabcolsep}{4pt}
\begin{tabular}{@{}lcccccc@{}}
\toprule
\textbf{Method} &
\parbox{1.1cm}{\centering\textbf{Non-Rect.}\\[-1pt]\textbf{Layouts}} &
\parbox{1.0cm}{\centering\textbf{Explicit}\\[-1pt]\textbf{Semantics}} &
\parbox{0.8cm}{\centering\textbf{Text}\\[-1pt]\textbf{Editing}} &
\parbox{1.0cm}{\centering\textbf{Trained}\\[-1pt]\textbf{Placement}} &
\parbox{1.0cm}{\centering\textbf{Asset}\\[-1pt]\textbf{Sampling}} \\
\midrule
ATISS \cite{paschalidou2021atiss}             & \textcolor{green}{\pmb{\checkmark}} & \textcolor{red}{\ding{55}} & \textcolor{red}{\ding{55}} & \textcolor{green}{\pmb{\checkmark}} & \textcolor{red}{\ding{55}} \\
Mi-Diff \cite{hu2024mixed}                    & \textcolor{green}{\pmb{\checkmark}} & \textcolor{red}{\ding{55}} & \textcolor{red}{\ding{55}} & \textcolor{green}{\pmb{\checkmark}} & \textcolor{red}{\ding{55}} \\
LayoutGPT \cite{feng2023layoutgpt}            & \textcolor{red}{\ding{55}} & \textcolor{green}{\pmb{\checkmark}} & \textcolor{red}{\ding{55}} & \textcolor{red}{\ding{55}} & \textcolor{red}{\ding{55}} \\
LayoutVLM \cite{sun2024layoutvlm}             & \textcolor{red}{\ding{55}} & \textcolor{green}{\pmb{\checkmark}} & \textcolor{red}{\ding{55}} & \textcolor{red}{\ding{55}} & \textcolor{red}{\ding{55}} \\
InstructScene \cite{lin2024instructscene}     & \textcolor{red}{\ding{55}} & \textcolor{red}{\ding{55}} & \textcolor{red}{\ding{55}} & \textcolor{green}{\pmb{\checkmark}} & \textcolor{red}{\ding{55}} \\
Ctrl-Room \cite{fang2025ctrl}                 & \textcolor{green}{\pmb{\checkmark}} & \textcolor{red}{\ding{55}} & \textcolor{red}{\ding{55}} & \textcolor{green}{\pmb{\checkmark}} & \textcolor{red}{\ding{55}} \\
SceneWeaver \cite{yang2025sceneweaver}        & \textcolor{red}{\ding{55}} & \textcolor{green}{\pmb{\checkmark}} & \textcolor{red}{\ding{55}} & \textcolor{green}{\pmb{\checkmark}} & \textcolor{red}{\ding{55}} \\

EditRoom \cite{zheng2025editroom}    & \textcolor{red}{\ding{55}} & \textcolor{red}{\ding{55}} & \textcolor{green}{\pmb{\checkmark}} & \textcolor{green}{\pmb{\checkmark}} & \textcolor{red}{\ding{55}} \\

DirectLayout \cite{ran2026direct}   & \textcolor{red}{\ding{55}} & \textcolor{green}{\pmb{\checkmark}} & \textcolor{red}{\ding{55}} & \textcolor{green}{\pmb{\checkmark}} & \textcolor{red}{\ding{55}} \\

\midrule
\textbf{ReSpace (ours)}                       & \textcolor{green}{\pmb{\checkmark}} & \textcolor{green}{\pmb{\checkmark}} & \textcolor{green}{\pmb{\checkmark}} & \textcolor{green}{\pmb{\checkmark}} & \textcolor{green}{\pmb{\checkmark}} \\
\bottomrule
\end{tabular}
\end{table}

\vspace{-3mm}

\paragraph{Language-based Scene Synthesis.} Early work like CLIP-Layout \cite{liu2023clip} explored text-prompted synthesis using CLIP \cite{radford2021learning} embeddings. With instruction-tuned LLMs, agent-based approaches evolved: LayoutGPT \cite{feng2023layoutgpt} pioneered zero-shot placement via CSS-based representation, while I-Design \cite{ccelen2024design}, Holodeck \cite{yang2024holodeck}, and Open-Universe \cite{aguina2024open} employ multi-agent systems to construct scene graphs or DSL instances before separate layout optimization. LayoutVLM \cite{sun2024layoutvlm} generates text-based layouts with constraints before optimization, LLPlace \cite{yang2024llplace} retrieves assets via text prompts before using a fine-tuned LLM for placement, and SceneCraft \cite{kumaran2023scenecraft} targets scene generation via iterative code generation with visual feedback. SceneWeaver \cite{yang2025sceneweaver} uses an LLM-based agent framework for text-driven scene synthesis. DirectLayout \cite{ran2026direct} generates numerical 3D layouts directly from text via LLM spatial reasoning with CoT-grounded rewards. RoomDreamer \cite{song2023roomdreamer} edits a scanned indoor mesh by generating text-guided coherent geometry and textures via diffusion and joint mesh optimization, without explicit object-level semantic layout modeling. CASAGPT \cite{feng2025casagpt} targets cuboid arrangement for interior design but lacks natural language semantics. More recently, Ctrl-Room \cite{fang2025ctrl} separates layout and appearance generation, achieving controllable text-to-3D room generation with mask-guided editing, with further recent advances in instruction-driven synthesis \cite{yang2026optiscene, bai2025freescene, choi2026scenenat}. However, recent methods either require separate optimization stages, focus on open-domain generation, or lack explicit semantics or editing capabilities. Table~\ref{tab:method_comparison} summarizes key properties across most similar recent methods. Unlike prior work, our approach uses a specialized trained LLM for indoor scene synthesis, directly predicting object semantics and positioning while supporting probabilistic asset sampling. This remains fully generative \cite{bucher2023performance} while extending beyond rectangular floor plans to non-convex geometries.

\vspace{-3mm}

\paragraph{Preference Alignment and Test-Time Compute Scaling.} LLM development has evolved from pre-training only (GPT-3 \cite{brown2020language} era) to dual-stage pipelines with instruction-tuning and preference alignment. Nominal works include InstructGPT \cite{ouyang2022training}, FLAN \cite{wei2021finetuned}, Reinforcement Learning from Human Feedback (RLHF) \cite{christiano2017deep}, Direct Preference Optimization (DPO) \cite{rafailov2023direct}, Rejection sampling Fine-Tuning (RFT) \cite{zelikman2022star, yuan2023scaling, dong2023raft}, and, most recently, Group Relative Policy Optimization (GRPO) \cite{shao2024deepseekmath} and RLVR \cite{lambert2024t, su2025crossing}. Recent work has also focused on increasing test-time compute \cite{snell2024scaling} via self-consistency \cite{wang2022self}, Best-of-N, and reward models \cite{brown2024large}. Formulating scene synthesis via language modeling, we employ preference alignment with verifiable rewards on this task.

\section{\textsc{ReSpace}}
\label{chap:method}
\begin{figure*}
    \centering
    \includegraphics[width=1.0\linewidth]{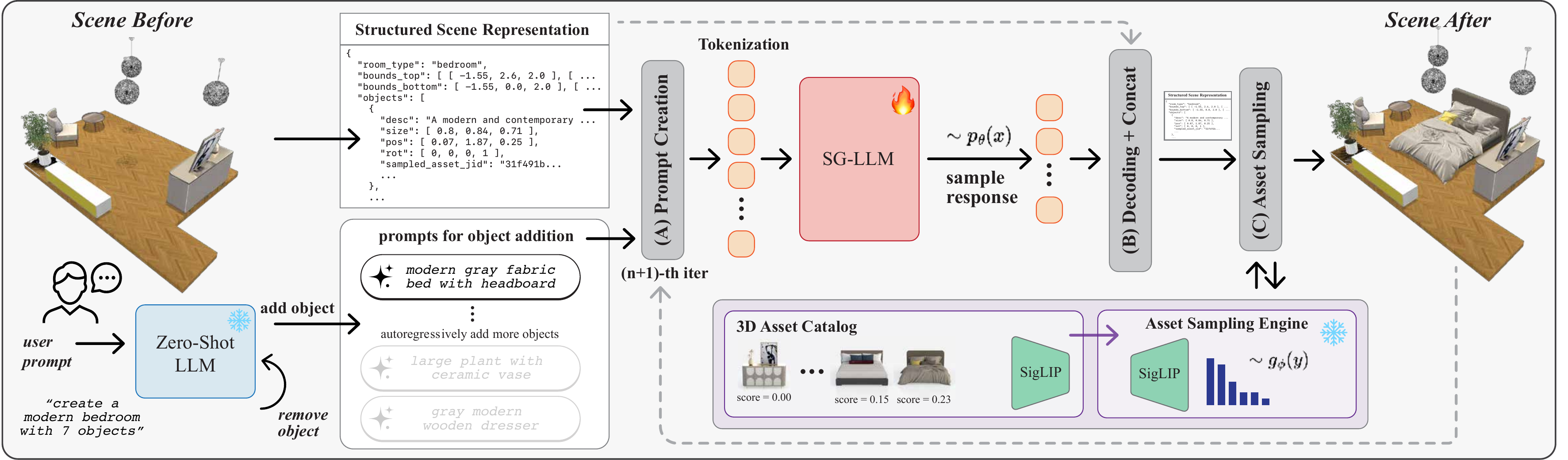}
    \vspace{-5mm}
    \caption{\textbf{ReSpace:} Given a text instruction and an existing scene in SSR, a zero-shot LLM emits sequential add/remove commands. Addition is done via our specially trained SG-LLM ($p_\theta$), removal via zero-shot SSR editing.}
    \label{fig:overview}
    \vspace{-3mm}
\end{figure*}

We introduce \textsc{ReSpace}, a method for autoregressive indoor scene generation and editing via natural language that sequentially adds and removes objects to empty or partial scenes (Fig. \ref{fig:overview}).

\vspace{1mm}

\textbf{Problem Statement.} Given a user instruction $u_i$ in natural language, our goal is to learn the conditional distribution $\hat{S}_i \thicksim p_{\theta}(S_i|u_i)$ with input scene $S_i$ and modified scene $\hat{S}_i$. Let $\mathcal{S} = \{S_1, S_2, ..., S_N\}$ be a collection of indoor scenes, where each scene $S_i = (T, \mathcal B, \mathcal O)$ is composed by its room type $T \in \mathcal{T}$, room boundaries $\mathcal{B} = \{\mathcal{B}_\text{top}, \mathcal{B}_\text{bottom}\}$, and unordered set of objects $\mathcal O = \{ O_1, O_2,..., O_K \}$. Unlike previous work, our bounds are defined as ordered point sets $b_i \in \mathbb{R}^{3}$ forming closed rectilinear polygons — $\mathcal{B}_\text{top} = \{ b_1, b_2, ... b_M\}$ for the ceiling and $\mathcal{B}_\text{bottom}$ for the floor. Further, each object in the scene $O_i = (d_i, h_i, t_i, r_i)$ is represented as a labeled 3D bounding box with asset description $d_i$, size $h_i \in \mathbb{R}^{3}$, position $t_i \in \mathbb{R}^3$, and orientation $r_i \in \mathbb{R}^{4}$. The object description $d_i$ captures fine-grained object semantics such as material, color, and style explicitly via text. Rotations are given as quaternions. We formulate the task of 3D scene synthesis as learning a generative model such that a scene with $K$ objects can be autoregressively composed from previously placed objects $\{ O_{j<i} \}$, natural language prompt $p_i$, room boundaries $\mathcal{B}$, and room type $T$.

\subsection{Structured Scene Representation}
\label{chap:ssr}
Given a scene $S_i = (T, \mathcal B, \mathcal O)$, we propose a Structured Scene Representation (SSR) that follows a nested dictionary schema. This is inspired by hierarchical DSLs as seen in prior work on neurosymbolic representations and shape programs \cite{zhang2024scene, tian2019learning, avetisyan2024scenescript}, as well as structured representations in Structured3D \cite{zheng2020structured3d} and SpatialLM \cite{mao2025spatiallm}, but follows a simpler structure for 3D indoor scenes. Let the room type be given as a short text string, let boundaries $\mathcal{B}_\text{top}$ and $\mathcal{B}_\text{bottom}$ be given as a nested list of 3D coordinates, and let the set of objects be a flat list, with each object defined as a dictionary with its compact textual description, size, position, and rotation. A full example of our SSR is given in \ref{supp:ssr-example} (Supp.), with a snippet also in Fig.\ref{fig:overview}. Note that the 3D asset choice is detached from the actual scene representation. Thus, SSR is an \textit{abstraction} over any scene instance and allows to swap assets without changing the underlying SSR. This choice, in contrast to neural scene or voxel-based methods \cite{peng2020convolutional, mildenhall2021nerf}, is lightweight ($\thicksim\text{KBs}$), and directly editable. Further, it is extensible, e.g., by representing doors/windows or adding spatial relationships between objects for more fine-grained scene graphs.

\subsection{Scene Synthesis via Autoregressive Modeling}
Given an SSR instance, we can tokenize a scene $S_i$ into $N$ text tokens $t_j$ such that $Tok(S_i) = \mathcal{U} = \{t_1,...,t_N\}$. Let $\mathcal{U}_{pr}$ be the sequence of tokens for the existing scene that composes an SSR, and let $\mathcal{U}_i$ be the token sequence for the current object. Let $p_i$ represent the object prompt for the next object to add. Note that $\text{Tok}(S_i) = \mathcal{U}_{\text{pr}} + \mathcal{U}_i$, where the complete scene tokenization is the concatenation of the existing scene tokens and the new object tokens. We can formulate a generative model for autoregressive scene synthesis and completion as a conditional next-token prediction task:

\vspace{-6mm}
\begin{equation}
\begin{aligned}
p_{\theta}\big( O_i | p_i, \{ O_{j<i} \}, T, \mathcal{B}\big) &= \prod_{j=0}^{M} p_{\theta}\big(t_j | p_i, \mathcal{U}_{pr}, t_{<j}\big)
\end{aligned}
\end{equation}
\vspace{-4mm}

\noindent thus, during inference, sampling the next object for the scene involves sampling $M$ tokens from $p_{\theta}$(x) until the end-of-sequence (EOS) token is chosen. Let, $p_{\theta}$(x) be represented by an LLM and let this specially trained model for autoregressive object addition be denoted as SG-LLM (Scene Graph LLM). We show our pipeline in Fig. \ref{fig:overview} for an example scene, where the full input string for SG-LLM is composed in step (A) from the existing SSR and a single object prompt. After tokenization, forward pass in the LLM, and response sampling, tokens get decoded and concatenated with the existing object list in step (B). Lastly, a 3D asset is sampled in step (C) via asset sampling engine. This process can be repeated $K$ times to iteratively add more objects, given $K$ object prompts. With object addition formulated as next-token prediction, SG-LLM is trained via supervised fine-tuning (SFT) and can be refined via RLVR (see \ref{supp:rlvr} in Supp.). 

\subsection{Stochastic Asset Sampling} 
We can retrieve assets for added objects from a given 3D asset catalog using the descriptions and sizes of each object defined in the SSR. Prior work uses greedy selection via closest 3D bounding box match, filtered by class label \cite{feng2023layoutgpt, paschalidou2021atiss, tang2024diffuscene, ccelen2024design, yang2024llplace, hu2024mixed}. In contrast, we formalize asset retrieval as a probabilistic process where each 3D asset mesh $m_i$ is drawn from a distribution parameterized by semantic and geometric constraints: $m_i \thicksim g_{\phi}(d_i, h_i)$, where $d_i$ is the natural language description and $h_i$ is the target size. The distribution $g_{\phi}$ computes scores as weighted combinations of semantic and geometric similarities: $\text{score}(m_j) = \lambda \cdot \text{sim}_{\text{sem}}(d_i, d_j) + (1-\lambda) \cdot \text{sim}_{\text{geo}}(h_i, h_j)$, where $\text{sim}_{\text{sem}}$ uses L2-normalized SigLIP embeddings for text-to-asset matching and $\text{sim}_{\text{geo}}$ measures size compatibility via Gaussian similarity: $\exp(-\|s_i - s_j\|^2/(2\sigma^2))$. The final distribution is obtained through temperature-scaled softmax with nucleus sampling (top-$p$) and top-$k$ filtering. For deterministic `greedy' retrieval, we can set $m_i = \text{argmax}_{m_j} g_{\phi}(d_i, h_i)$.


\begin{figure}
    \centering
    \includegraphics[width=1.0\linewidth]{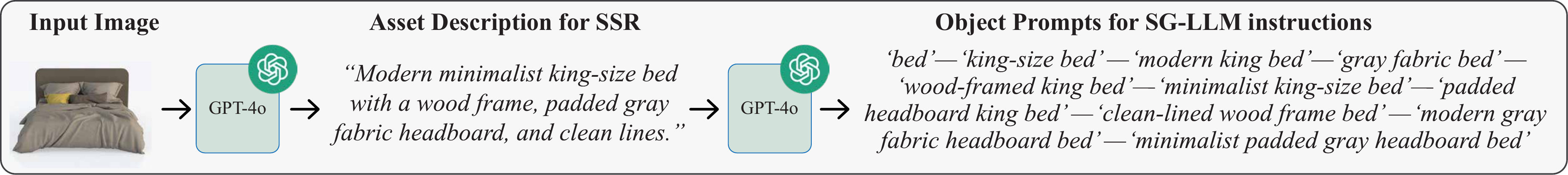}
    \vspace{-6mm}
    \caption{Example of description and prompt bank for an asset.}
    \label{fig:prompt-desc-example}
    \vspace{-5mm}
\end{figure}

\subsection{Removal/Full Scenes via Zero-Shot Learning} 
Our method enables scene \textit{editing} via autoregressive addition and removal using a zero-shot LLM. For removal, this LLM directly modifies the SSR JSON. For full scene generation, we leverage the LLM's inherent \textit{world model} to generate object prompt lists $\mathcal{P}_i = \{p_{1},...,p_{K}\} \thicksim LLM_{ZS}(u_i)$ from user instruction $u_i$, which SG-LLM processes autoregressively. While a unified model would be preferable, we deliberately specialize SG-LLM for object addition, the sub-task where frontier zero-shot reasoning fails (Sec.~\ref{chap:experiments}), while delegating removal and instruction decomposition to a zero-shot LLM, where it already suffices. This division focuses trained capacity where it is needed and avoids mode collapse from task/class imbalance. System prompts are in \ref{supp:prompts-zero-shot} (Supp.).

\subsection{Voxelization-Based Loss for Layout Violations}
\label{chap:lay-viol}
\begin{figure}
    \centering
    \scriptsize
    \setlength{\tabcolsep}{2pt}
    \begin{tabular}{cc}
        \includegraphics[width=0.46\linewidth]{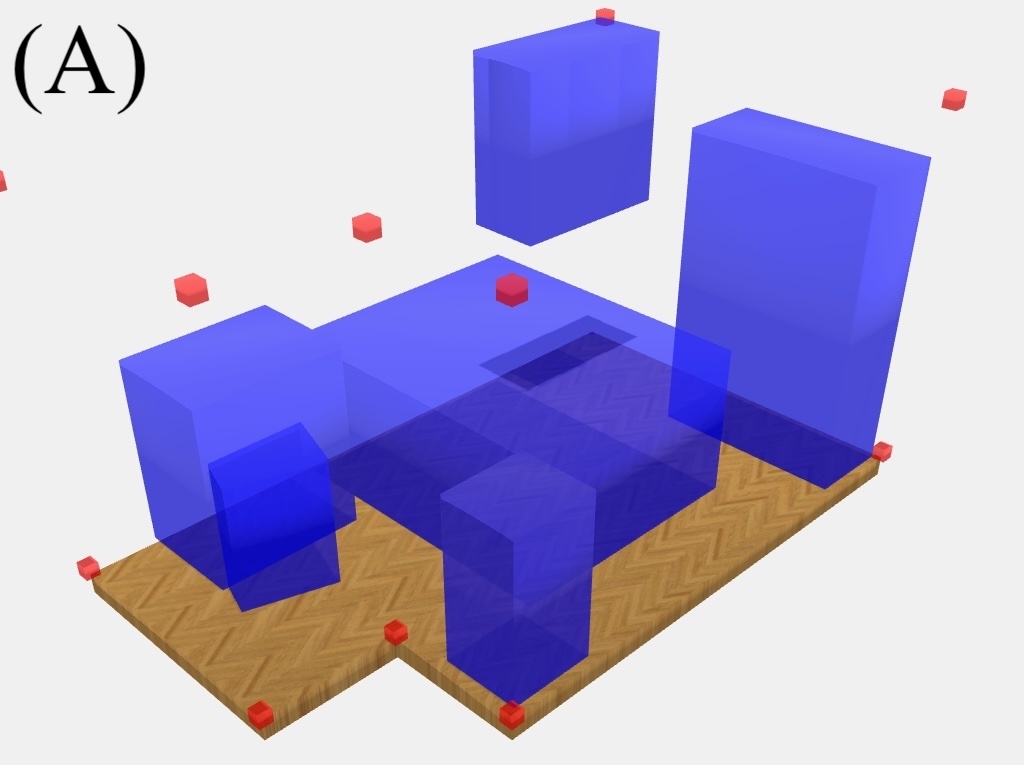} &
        \includegraphics[width=0.46\linewidth]{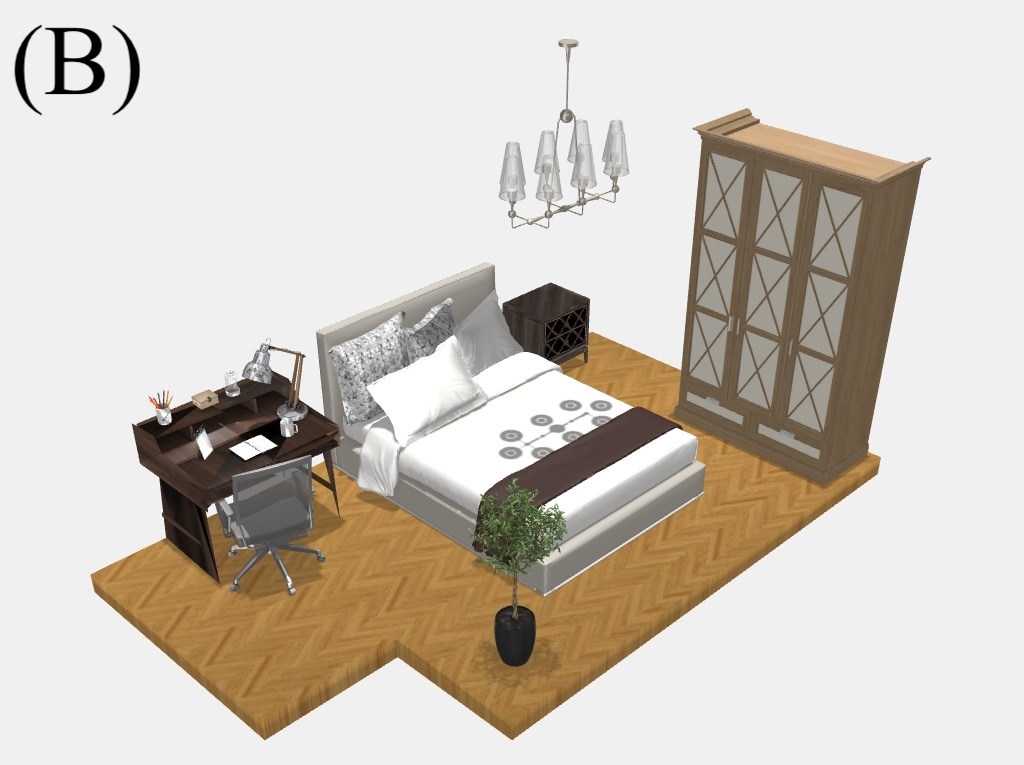} \\
        \includegraphics[width=0.46\linewidth]{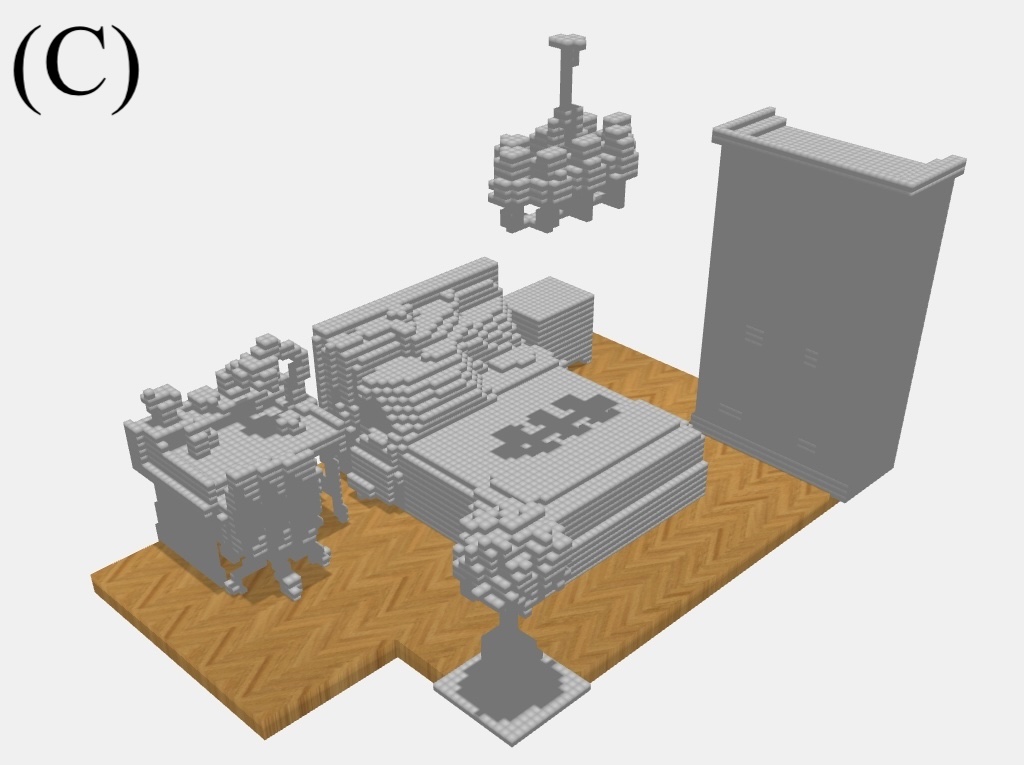} &
        \includegraphics[width=0.34\linewidth]{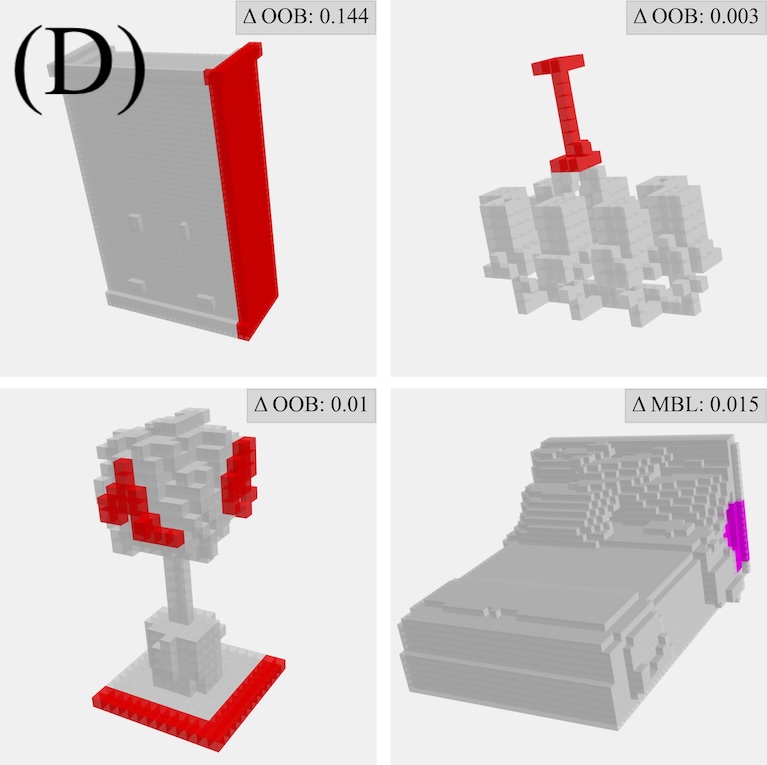} \\
    \end{tabular}
    \vspace{-2mm}
    \caption{%
        \textbf{(A)} Scene with 3D bounding boxes in blue and bounds in red. \textbf{(B)} Scene with 3D meshes. \textbf{(C)} Scene with voxels. \textbf{(D)} Examples of OOB/MBL voxel violations. Note how desk/chair interact smoothly in mesh space compared to their bounding boxes, while the lamp is largely OOB as box but only minor as mesh.
    }
    \label{fig:lay-viol}
    \vspace{-3mm}
\end{figure}

Representing scenes via SSR follows previous work in that objects are simplified as a collection of positioned 3D bounding boxes, visualized in Fig.~\ref{fig:lay-viol} (A) with blue boxes for objects and red cubes for ceiling and floor room bounds. Existing work on indoor scene synthesis does not study layout violations extensively --- reporting only the ratio of scenes with objects partially out-of-bounds \cite{ccelen2024design}, or object intersections via Intersection-over-Union \cite{hu2024mixed} or average volume intersection \cite{ccelen2024design} on 3D bounding boxes. However, bounding-box-based metrics cannot accurately evaluate realistic object placement (e.g., chair partially under table in Fig.~\ref{fig:lay-viol} (A)) or provide fine-grained violation signals.

We introduce the Voxelization-Based Loss (VBL), a geometry-aware \textit{evaluation metric} defined as follows. We voxelize the scene boundary mesh $\mathcal{B}$ with fixed voxel size $G$ to create a uniform grid $V_S$, and similarly voxelize each object mesh $O_j$ into a binary occupancy grid $\mathcal{V}_j \in \{0,1\}^{x_j \times y_j \times z_j}$, where each voxel indicates whether that spatial location is occupied by the object. To quantify layout violations, we define two complementary sub-metrics: (1) Out-of-Bounds Loss (OOB) counts voxels outside scene boundaries as $\text{OOB}_j = \sum_i \mathcal{V}_j(i) - \sum_i \mathcal{V}_j(i) \cdot \mathcal{V}_S(i)$, with total $\text{OOB} = \sum_j \text{OOB}_j$, and (2) Mesh Boundary Loss (MBL) measures voxel overlap between unique object pairs $(O_m, O_n)$ as $\text{MBL}_{(m,n)} = \sum_i \mathcal{V}_m(i) \cdot \mathcal{V}_n(i)$, computed once per unique pair with total $\text{MBL} = \sum_{m < n} \text{MBL}_{(m,n)}$. The complete VBL is the sum: $\text{VBL} = \text{OOB} + \text{MBL}$. OOB and MBL capture orthogonal failure modes --- objects outside boundaries have high OOB but low MBL since they rarely intersect with other objects --- making it crucial to minimize both. Fig.~\ref{fig:lay-viol} visualizes these violations with OOB voxels in red and MBL voxels in purple. Since MBL scales subquadratically with object count, we implement a horizontal 2D intersection check for early stopping, skipping full 3D voxel computation when 2D projections of object pairs show zero overlap, significantly reducing computation time for scenes with many spatially distant objects. We empirically find that a voxel size of $G=0.05m$ provides an optimal trade-off between computational efficiency and accuracy.

\section{Experiments}
\label{chap:experiments}
We conduct experiments on four tasks: (i) single object addition on partial scenes via prompt-based instructions; (ii) object removal via zero-shot LLM; (iii) autoregressive editing sequences chaining additions and removals to simulate realistic multi-step user interactions; and (iv) full scene synthesis as an application, enabling comparison with a broader set of baselines. We further perform ablation studies on scene complexity, prompt complexity, test-time compute scaling via Best-of-N (BoN) sampling, and asset-agnostic spatial reasoning.

\vspace{-3mm}

\paragraph{Scene-Prompt Dataset.} We partition 3D-FRONT \cite{fu20213d_front} into \texttt{`bed'} (bedrooms), \texttt{`liv'} (living rooms and dining rooms), and \texttt{`all'} splits (with 6328/500/500, 3830/500/500, and 12055/500/500 train/val/test samples respectively) after filtering out noisy samples via our voxelization-based method. Since the dataset lacks textual descriptions, we use GPT-4o \cite{hurst2024gpt} as a vision-language model to generate detailed descriptions $\bf{d}_j$ for each object from provided 3D-FUTURE \cite{fu20213d_future} renderings, following vision-based approaches for labeling \cite{raghu2023towards, aguina2024open}. We then create 10 unique, concise prompts per object to form our prompt bank $\mathcal{P}(o)$ (see Fig. \ref{fig:prompt-desc-example}). The prompt bank provides data augmentation with varying levels of detail for the same object—for example, a bed might have prompts ranging from ``bed" (one word) to ``modern king-size platform bed" (four words). During training, we sample $p_i \sim \text{Unif}(\mathcal{P}(o))$ for each object, ensuring the model learns a robust prompt-to-object distribution across different prompt styles. During training, we dynamically generate instruction triples, and include empty rooms (10\%), scenes with a final object placement missing (10\%), and partial scenes with random number of existing objects (80\%). We create fixed test sets with 500 instructions and corresponding object prompts using three random seeds. More details are outlined in \ref{supp:preprocessing} (Supp.).

\begin{figure}
    \centering
    \includegraphics[width=1.0\linewidth]{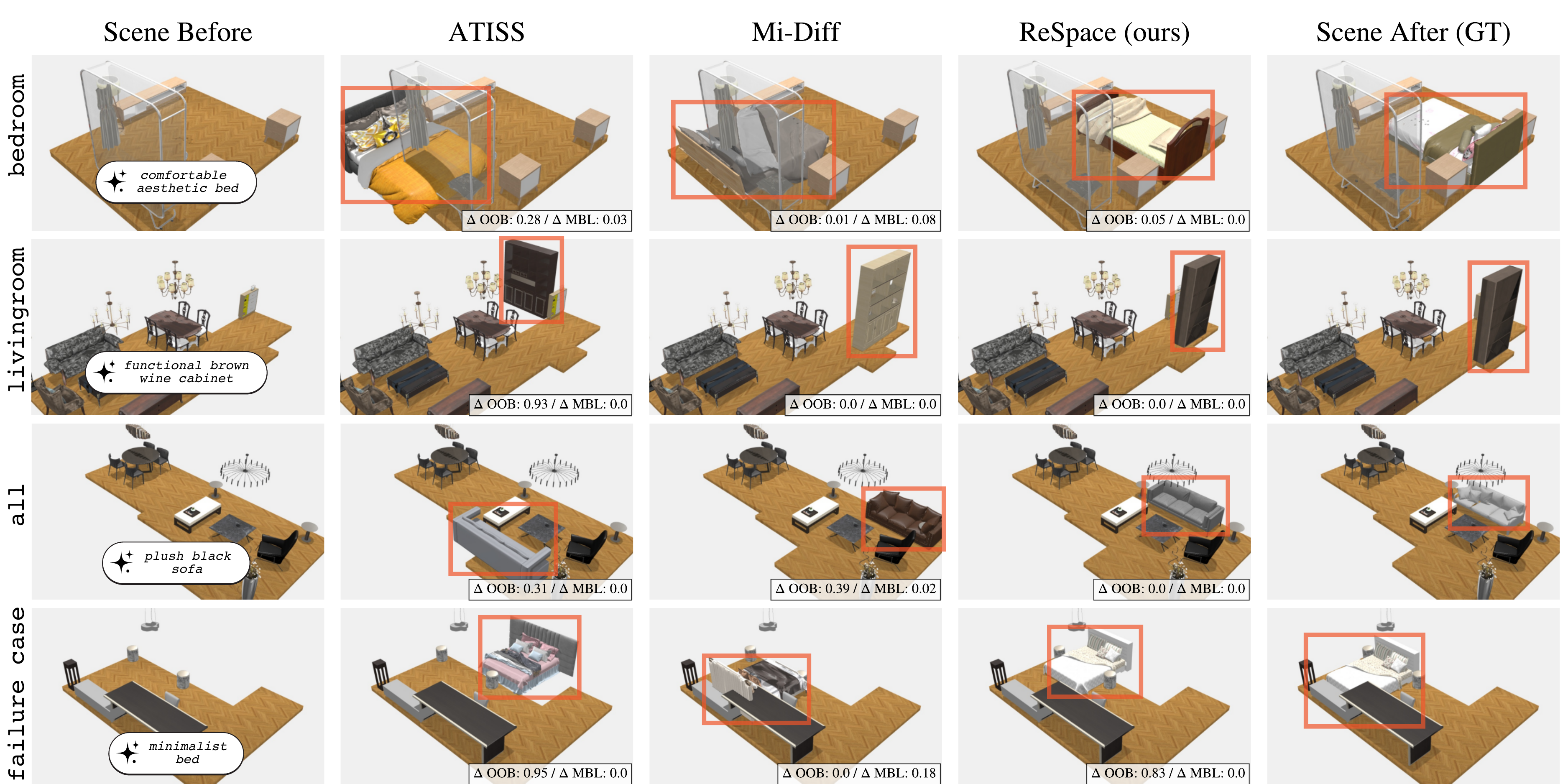}
    \vspace{-5mm}
    \caption{Qualitative results on single instructions, with our method performing the strongest. For ours, we use $\text{ReSpace/A}^{\dagger}$. We show a failure case on the last row.}
    \label{fig:instr-qualitative-samples}
    \vspace{-4mm}
\end{figure}

\subsection{Experimental Settings}
\label{chap:exp-settings}

\paragraph{Baselines and Implementation.} We compare against trained 3D indoor scene synthesis baselines: (i) ATISS \cite{paschalidou2021atiss}, a transformer-based auto-regressive model; and (ii) MiDiffusion \cite{hu2024mixed} (Mi-Diff), a mixed discrete-continuous diffusion model. Both take a $256{\times}256$ top-down orthographic projection of the floor plan as input condition. ATISS is natively autoregressive and supports single object-addition via one-hot class label conditioning; for Mi-Diff, we follow their masking strategy and enable de-noising for a single object only. \textbf{We select baselines on a single criterion: support for floor-plan--conditioned object placement, the regime our task and VBL metric require.} ATISS and Mi-Diff condition on arbitrary (incl.\ non-rectangular) floor plans and serve as our primary baselines; LayoutGPT \cite{feng2023layoutgpt} and LayoutVLM \cite{sun2024layoutvlm} support only rectangular plans and are evaluated on a rectangular-only subset for full scene synthesis. We exclude InstructScene, EditRoom, OptiScene, and DirectLayout as none conditions on an explicit (non-rectangular) floor-plan boundary, making a VBL-based comparison ill-defined. Note that LayoutVLM requires a predefined set of objects with bounding boxes as input, whereas our method generates the object list from scratch. For fair comparison, we use greedy asset selection, as baselines only support deterministic retrieval. We define a high-quality placement filter requiring $\text{VBL} < 10^{-5}$, $\text{PMS} > 0.85$, and relative size L2 $< 0.5$, used both as a binary reward signal for RLVR training of SG-LLM and as accuracy criterion for evaluation on editing sequences. Full implementation details, including prompt-to-class-label mapping, are in \ref{supp:impl-details}, details on stochastic assets are in \ref{supp:asset-sampling} (Supp.).

\begin{table*}
  \caption{Quantitative evaluation on \textbf{single object addition} over a hold-out test set of $3 \times 500$ instructions with 3 random seeds. KID and layout violations (OOB, MBL, VBL) are scaled by $10^3$ for readability. Best values are \textbf{bold}, second best \underline{underlined}.}
  \vspace{-2mm}
  \label{exps-scenes-instr}
  \centering
  \scriptsize
  \resizebox{\textwidth}{!}{%
  \begin{tabular}{@{}c l*{7}{r}@{}}
    \toprule
    & & \multicolumn{3}{c}{Layout Violations} & \multicolumn{3}{c}{Scene Renderings} & \multicolumn{1}{c}{Prompting} \\
    \cmidrule(r){3-5} \cmidrule(r){6-8} \cmidrule(r){9-9}
    & Method & \multicolumn{1}{c}{$\text{OOB}^{\Delta}_{\times1e3} \downarrow$} & \multicolumn{1}{c}{$\text{MBL}^{\Delta}_{\times1e3} \downarrow$} & \multicolumn{1}{c}{$\text{VBL}^{\Delta}_{\times1e3} \downarrow$} & \multicolumn{1}{c}{$\text{FID} \downarrow$} & \multicolumn{1}{c}{$\text{FID}_\text{CLIP} \downarrow$} & \multicolumn{1}{c}{$\text{KID}_{\times\text{1e3}} \downarrow$} & \multicolumn{1}{c}{$\text{PMS} \uparrow$} \\
    
    \midrule
    
    \multirow{4}{*}{\rotatebox[origin=c]{90}{\texttt{`bed'}}} 
    
    & ATISS & $97.70_{\pm 6.0}$ & $13.54_{\pm 0.5}$ & $111.24_{\pm 5.4}$ & $36.18_{\pm .3}$ & $1.74_{\pm .0}$ & $0.19_{\pm .0}$ & $0.58_{\pm .0}$ \\
    
    & Mi-Diff & $64.04_{\pm 5.3}$ & $14.27_{\pm 1.5}$ & $78.31_{\pm 4.1}$ & $36.12_{\pm .3}$ & $1.76_{\pm .1}$ & ${0.05}_{\pm .0}$ & $0.57_{\pm .0}$ \\
    
    & $\text{ReSpace/B}$
    & $\underline{11.77}_{\pm 3.7}$ 
    & $\underline{4.45}_{\pm 0.5}$ 
    & $\underline{16.23}_{\pm 4.0}$ 
    & $\mathbf{35.23}_{\pm .3}$ 
    & $\mathbf{1.64}_{\pm .0}$ 
    & $\mathbf{-0.06}_{\pm .0}$ 
    & $\underline{0.88}_{\pm .0}$ \\
    
    & $\text{ReSpace/A}^{\dagger}$ 
    & $\mathbf{10.75}_{\pm 2.6}$ 
    & $\mathbf{3.91}_{\pm 0.7}$ 
    & $\mathbf{14.66}_{\pm 2.4}$ 
    & $\underline{35.35}_{\pm .2}$ 
    & $\underline{1.66}_{\pm .0}$ 
    & $\underline{-0.03}_{\pm .1}$ 
    & $\mathbf{0.89}_{\pm .0}$ \\
    
    \midrule
    
    \multirow{4}{*}{\rotatebox[origin=c]{90}{\texttt{`liv'}}} 
    
    & ATISS & $63.87_{\pm 6.9}$ & $11.43_{\pm 3.8}$ & $75.30_{\pm 5.8}$ & $32.26_{\pm .1}$ & $1.48_{\pm .0}$ & $\underline{0.71}_{\pm .3}$ & $0.58_{\pm .0}$ \\
    
    & Mi-Diff & $43.88_{\pm 7.6}$ & $12.87_{\pm 1.4}$ & $56.75_{\pm 8.8}$ & $33.30_{\pm .3}$ & $1.53_{\pm .0}$ & $1.06_{\pm .2}$ & $0.56_{\pm .0}$ \\
    
    & $\text{ReSpace/L}$
    & $\underline{10.68}_{\pm 0.5}$ 
    & $\underline{4.27}_{\pm 0.3}$ 
    & $\underline{14.95}_{\pm 0.7}$ 
    & $\underline{31.94}_{\pm .0}$ 
    & $\mathbf{1.40}_{\pm .0}$ 
    & $\mathbf{0.21}_{\pm .1}$ 
    & $\underline{0.84}_{\pm .0}$ \\
    
    & $\text{ReSpace/A}^{\dagger}$ 
    & $\mathbf{7.51}_{\pm 1.7}$ 
    & $\mathbf{3.11}_{\pm 0.9}$ 
    & $\mathbf{10.62}_{\pm 2.5}$
    & $\mathbf{31.84}_{\pm .0}$ 
    & $\underline{1.41}_{\pm .0}$ 
    & $1.41_{\pm .0}$ 
    & $\mathbf{0.87}_{\pm .0}$ \\
    
    \midrule
    
    \multirow{3}{*}{\rotatebox[origin=c]{90}{\texttt{`all'}}} 
    
    & ATISS & $121.66_{\pm 8.6}$ & $\underline{14.48}_{\pm 1.0}$ & $136.14_{\pm 8.7}$ & $36.40_{\pm .0}$ & $1.77_{\pm .0}$ & $0.22_{\pm .1}$ & $\underline{0.57}_{\pm .0}$ \\
    
    & Mi-Diff & $\underline{40.51}_{\pm 5.5}$ & $18.19_{\pm 0.6}$ & $\underline{58.70}_{\pm 4.9}$ & $\underline{36.14}_{\pm .2}$ & $\underline{1.72}_{\pm .0}$ & $\underline{0.07}_{\pm .1}$ & $0.56_{\pm .0}$ \\

    
    & $\text{ReSpace/A}^{\dagger}$ 
    & $\mathbf{7.61}_{\pm 1.8}$ 
    & $\mathbf{3.60}_{\pm 1.0}$ 
    & $\mathbf{11.21}_{\pm 2.3}$ 
    & $\mathbf{35.41}_{\pm .3}$ 
    & $\mathbf{1.66}_{\pm .0}$ 
    & $\mathbf{-0.06}_{\pm .1}$ 
    & $\mathbf{0.87}_{\pm .0}$ \\
    
    \bottomrule
  \end{tabular}%
  }
\end{table*}

\vspace{-3mm}

\paragraph{Evaluation Metrics.} We use our introduced Voxelization-Based Loss (VBL) (see Section \ref{chap:lay-viol}) as the main evaluation metric, and follow previous work \cite{paschalidou2021atiss, lin2024instructscene, hu2024mixed} by rendering a top-down projection for each scene, computing Fréchet Inception Distance (FID) \cite{heusel2017gans}, $\text{FID}_{\text{CLIP}}$ \cite{kynkaanniemirole}, and Kernel Inception Distance (KID) \cite{binkowski2018demystifying} between the train split and generated scenes. For train split scenes, we compute a set of $\text{min}(N, 5000)$ renderings for instructions (partial scenes) and full scenes respectively. We also report Prompt Matching Score (PMS) to measure how many words $w_j$ from the prompt $p_i$ are captured via the description $d_i$ of the \textit{sampled} 3D asset: $\text{PMS}(p_i, d_i) = \frac{1}{|p_i|} \sum_{w_j \in p_i} \mathbbm{1}_{w_j \in d_i}$, with higher recall indicating better instruction-following capabilities. We use the postfixes `/B', `/L', and `/A' to denote the training room split for SG-LLM, and denote with $\text{ReSpace/A}^{\dagger}$ the model with additional preference alignment (details in \ref{supp:rlvr}).

\vspace{1mm}

\subsection{Prompt-Driven Scene Editing and Synthesis}

\paragraph{Object Addition.}
We present results for object addition in Table \ref{exps-scenes-instr}, reporting \textit{delta} VBL to quantify layout changes after insertion. Our method consistently outperforms baselines across all metrics and datasets. This advantage is not limited to trained baselines: fine-grained placement is also \emph{not} solved by frontier zero-shot prompting, where replacing SG-LLM with GPT-5.4-mini under identical SSR prompting yields $\text{VBL}^{\Delta}{=}39.78$ vs.\ $11.21$ on \texttt{`all'} (${\sim}3.5{\times}$ worse) and $\text{PMS}{=}0.57$ vs.\ $0.87$, despite the frontier model being orders of magnitude larger than our $1.5$B SG-LLM --- scale does not substitute for task-specific spatial reasoning. The model trained on \texttt{`all'} performs stronger even on \texttt{`bed'} and \texttt{`liv'} subsets, indicating that diverse training scenes help generalization. We further study preference alignment with verifiable rewards as an empirical exploration, finding it prone to reward hacking, e.g., shrinking object sizes to trivially lower VBL (\ref{supp:rlvr}). While GRPO improves on OOB, it does not consistently outperform SFT on MBL, and does not yield a meaningful improvement in human-perceived scene quality. DPO achieves the strongest layout violation metrics, yet this does not translate to human preference, likely due to overfitting to single-object placement rewards at the cost of broader scene coherence. RFT is the most stable across automatic metrics and human evaluation, with a modest but consistent preference over SFT ($58.8\%$) and a win over DPO; we adopt it for our final model $\text{ReSpace/A}^{\dagger}$, with full comparison in \ref{supp:user-study-1}. Fig.~\ref{fig:instr-qualitative-samples} shows qualitative results for single object additions, with our method sometimes exceeding ground-truth placements (e.g., sofa in third row). The last row shows a failure case with out-of-bounds placement. Beyond geometry, our strong OOB margin validates explicit boundaries over fixed-resolution floor plans for non-rectangular layouts, while PMS, our text-faithfulness measure, separates ReSpace from all baselines (${\geq}0.87$ vs ${\leq}0.57$), evidencing better prompt adherence beyond geometric validity.

\vspace{-3mm}

\paragraph{Object Removal.}
We additionally experiment with object removal using the same test instructions, re-merging the intended (ground-truth) object into the scene and prompting the system to \textit{remove} it given solely the object prompt. Since duplicate assets share the same description $d_i$, a removal is counted correct only if all assets matching the prompt are removed 1:1; we report accuracy as $(\text{\# correct} / \text{\# all})$, with $90.9\% \pm 0.6$ on \texttt{`bed'}, $75.2\% \pm 1.0$ on \texttt{`liv'}, and $87.3\% \pm 0.7$ on \texttt{`all'} using a zero-shot Llama-3.1-8B. Accuracy drops with SSR length (95\% at $<$200 words to $<$35\% at $>$500 words), reflecting the 8B model's long-context limits rather than semantic ambiguity: swapping in a stronger frontier model (GPT-5.4-mini) raises removal to $99.8\% \pm 0.2$ with no systematic failure pattern (see \ref{supp:removal-analysis}). Removal, an identification-and-deletion task, thus scales with the zero-shot capabilities of existing LLMs and needs no training.

\begin{figure}
    \includegraphics[width=\linewidth]{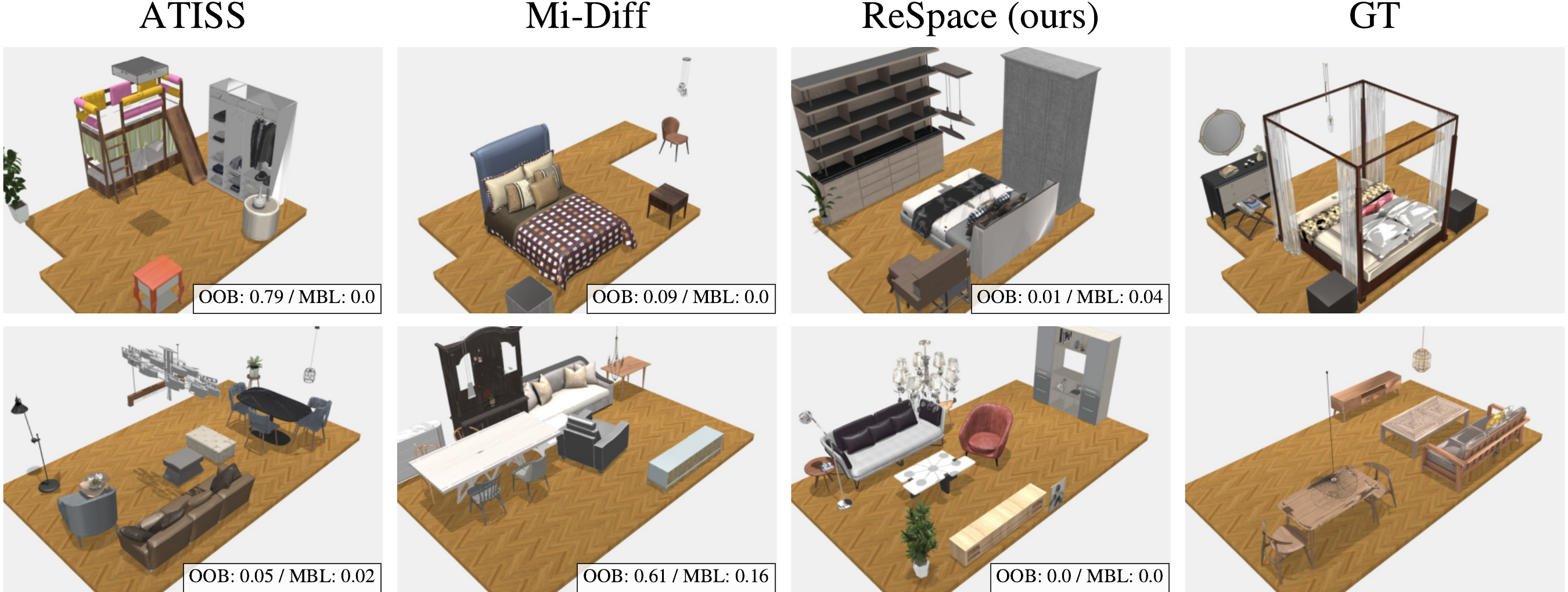}
    \vspace{-6mm}
    \caption{Qualitative results on full scenes (with $\text{ReSpace/A}^{\dagger}$).}
    \label{fig:full-qualitative-samples}
    \vspace{-5mm}
\end{figure}

\vspace{-3mm}

\paragraph{Autoregressive Editing Sequences.}
We evaluate ReSpace on synthetic autoregressive editing sequences of 1--10 instructions to simulate realistic user interactions, where a user iteratively adds and removes objects from a scene. Sequences are generated via random orderings of additions ($p{=}0.8$) and removals ($p{=}0.2$) applied to $3{\times}200$ test scenes. Per-step accuracy for addition uses our high-quality placement filter defined in Section~\ref{chap:exp-settings}, noting that it does not account for valid placements deviating significantly in bounding box size from the ground truth. For removals we use the same criterion as in Section~\ref{chap:full-scene-synth}. Fig.~\ref{fig:analysis-grid} (Top-left) shows per-step accuracy binned by sequence length for $\text{BoN}{=}\{1,8\}$. Ground-truth additions pose an imperfect upper bound as they do not always satisfy the strict filter either. The results show that with a single trial for additions, accuracy decreases with longer sequences as placement errors compound. However, increasing test-time compute shows consistently higher accuracy if we allow for more trials ($\text{BoN}{=}8$) and/or include rotation augmentation (rotate x4, pick best).

\vspace{-3mm}

\paragraph{Full Scene Synthesis.}
\label{chap:full-scene-synth}
Beyond object-level editing, we evaluate the application of full scene synthesis, to compare our work with this objective. Unlike end-to-end trained baselines, our method relies on prompts from a zero-shot LLM informed with: (1) 3D-FRONT object classes, (2) floor area to object count priors, and (3) few-shot prompt examples. While Table~\ref{exps-scenes-full} (non-rect.; Supp.) and Table~\ref{exps-scenes-full-rect-only} show higher FID/KID scores for ours, indicating a slightly larger gap to the original training distribution, ReSpace achieves substantially lower layout violation metrics. A current limitation is that object counts are fixed by the zero-shot LLM prior to placement and not adapted afterward, and we do not filter for over-population, so a room can be overfilled relative to its floor area; in our target interactive setting, a human-in-the-loop can guide or adjust the outcome.

To validate our approach against all baselines including those restricted to rectangular layouts (LayoutGPT, LayoutVLM), we conducted a comprehensive human evaluation study on a rectangular-only subset using $\text{ReSpace/A}^{\dagger}$ with $B_1{+}R{+}S_8$ (see \ref{chap:ablation} (iii)). With 334 participants performing 10,307 pairwise comparisons across 100 scenes, \textbf{ReSpace achieves a win rate of 75.3\%} --- more than 18 percentage points above second-placed Mi-Diff (56.7\%). While ReSpace achieves substantially lower layout violation metrics, FID/KID scores are slightly higher or on par with baselines, reflecting a distribution shift from training data. This confirms that closing the gap to the training distribution is not sufficient on its own; the user study suggests that human preference reflects a balance between layout quality and proximity to a reference distribution of indoor scenes, validating our design choices and highlighting the importance of human evaluation as a complementary signal to metrics alone. Details on the study and Bradley-Terry rankings are in \ref{supp:user-study-3}. Numerical results are in Tab.~\ref{exps-scenes-full-rect-only}. Fig.~\ref{fig:full-qualitative-samples} and \ref{fig:full-qualitative-samples-supp} show qualitative outputs. Lastly, runtime analysis shows that ReSpace achieves competitive runtime while supporting richer capabilities than all baselines (see \ref{supp:runtime-analysis} for details).

\begin{table*}
\caption{Quantitative evaluation on \textbf{rectangular-only scenes} from the \texttt{`all'} split with subset of $3 \times 257$ scenes. Metrics follow Table \ref{exps-scenes-instr}.}
  \vspace{-2mm}
  \label{exps-scenes-full-rect-only}
  \scriptsize
  \resizebox{\textwidth}{!}{%
  \begin{tabular}{@{}c l*{7}{r}@{}}
    \toprule
    & & \multicolumn{3}{c}{Layout Violations} & \multicolumn{3}{c}{Scene Renderings} & \multicolumn{1}{c}{Prompt} \\
    \cmidrule(r){3-5} \cmidrule(r){6-8} \cmidrule(r){9-9}
    & Method & \multicolumn{1}{c}{$\text{OOB}_{\times\text{1e3}} \downarrow$} & \multicolumn{1}{c}{$\text{MBL}_{\times\text{1e3}} \downarrow$} & \multicolumn{1}{c}{$\text{VBL}_{\times\text{1e3}} \downarrow$} & \multicolumn{1}{c}{$\text{FID} \downarrow$} & \multicolumn{1}{c}{$\text{FID}_\text{CLIP} \downarrow$} & \multicolumn{1}{c}{$\text{KID}_{\times\text{1e3}} \downarrow$} & \multicolumn{1}{c}{$\text{PMS} \uparrow$} \\
    \midrule
    
    \multirow{6}{*}{\rotatebox[origin=c]{90}{\texttt{`rect'}}}

     & $\text{LayoutGPT}$ 
    & $1199.7_{\pm 57.6}$ 
    & $84.2_{\pm 06.0}$ 
    & $1284.0_{\pm 63.3}$ 
    & $106.75_{\pm .5}$ 
    & $9.17_{\pm .1}$ 
    & $38.97_{\pm 1.}$ 
    & $n/a$ \\
    
    & $\text{ATISS}$
    & $403.8_{\pm 03.9}$ 
    & $88.1_{\pm 06.4}$ 
    & $491.9_{\pm 07.9}$ 
    & $70.69_{\pm .8}$ 
    & $\underline{4.22}_{\pm .0}$ 
    & $2.71_{\pm.7}$ 
    & $n/a$ \\
    
    & $\text{Mi-Diff}$
    & $236.0_{\pm 31.7}$ 
    & $74.9_{\pm 07.2}$ 
    & $310.8_{\pm 26.4}$ 
    & $\underline{69.75}_{\pm .6}$ 
    & $\mathbf{4.06}_{\pm .1}$ 
    & $\mathbf{1.37}_{\pm .5}$ 
    & $n/a$ \\
    
    & $\text{LayoutVLM}$ 
    & $78.6_{\pm 02.2}$ 
    & $84.3_{\pm 03.6}$ 
    & $162.9_{\pm 05.2}$ 
    & $80.04_{\pm .6}$ 
    & $5.91_{\pm .1}$ 
    & $6.33_{\pm .4}$ 
    & $n/a$ \\
    
    & $\text{ReSpace/A}^{\dagger}$ 
    & $\underline{70.5}_{\pm 10.9}$ 
    & $\underline{66.6}_{\pm 06.2}$ 
    & $\underline{137.1}_{\pm 17.1}$ 
    & ${70.15}_{\pm .4}$ 
    & $4.38_{\pm .1}$ 
    & $\underline{1.62}_{\pm .2}$ 
    & $\underline{0.71}_{\pm .0}$ \\
    

    & $\text{ReSpace/A}^{\dagger}_{S8{+}R}$
    & $\mathbf{4.6}_{\pm 00.6}$ 
    & $\mathbf{11.2}_{\pm 06.1}$ 
    & $\mathbf{15.8}_{\pm 06.8}$ 
    & $\mathbf{69.24}_{\pm 1.}$ 
    & $4.34_{\pm .1}$ 
    & ${1.96}_{\pm .4}$
    & $\mathbf{0.90}_{\pm .0}$ \\
    
    \bottomrule
  \end{tabular}%
  }
\vspace{-2mm}
\end{table*}

\subsection{Discussion}
\label{chap:ablation}

\begin{figure}
    \centering
    \scriptsize
    \setlength{\tabcolsep}{2pt}
    \begin{tabular}{cc}
        \includegraphics[width=0.46\linewidth]{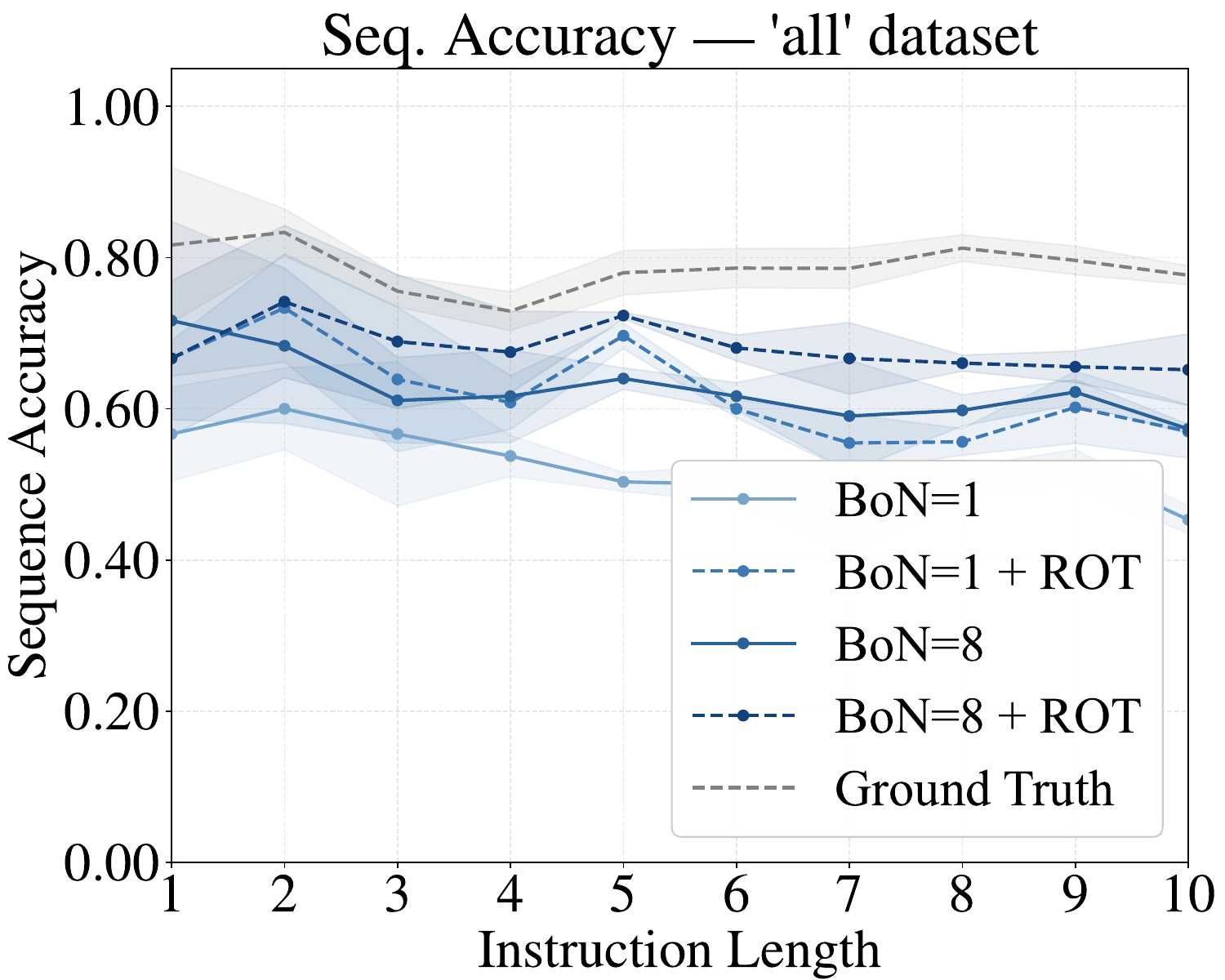} &
        \includegraphics[width=0.46\linewidth]{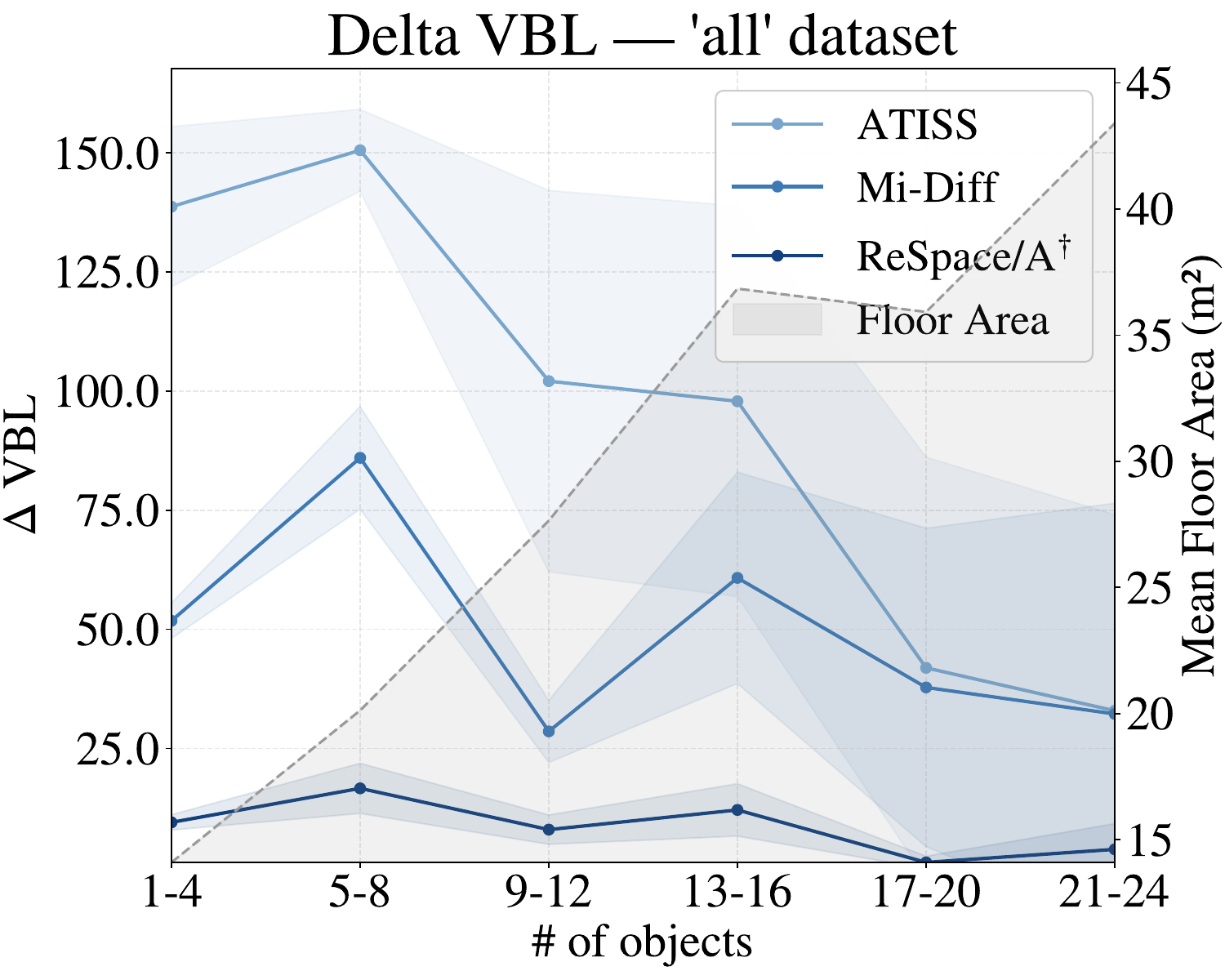} \\
        \includegraphics[width=0.46\linewidth]{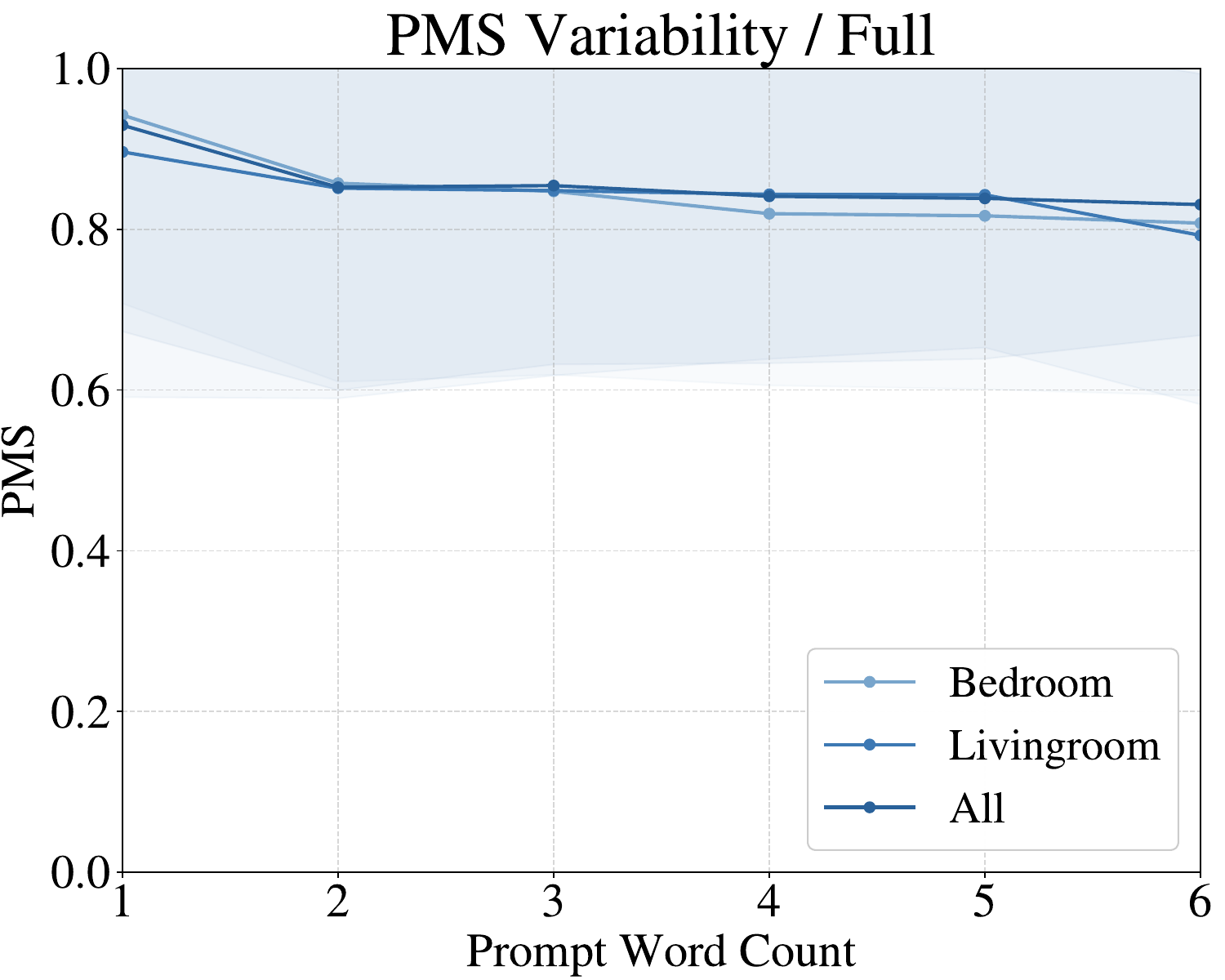} &
        \includegraphics[width=0.46\linewidth]{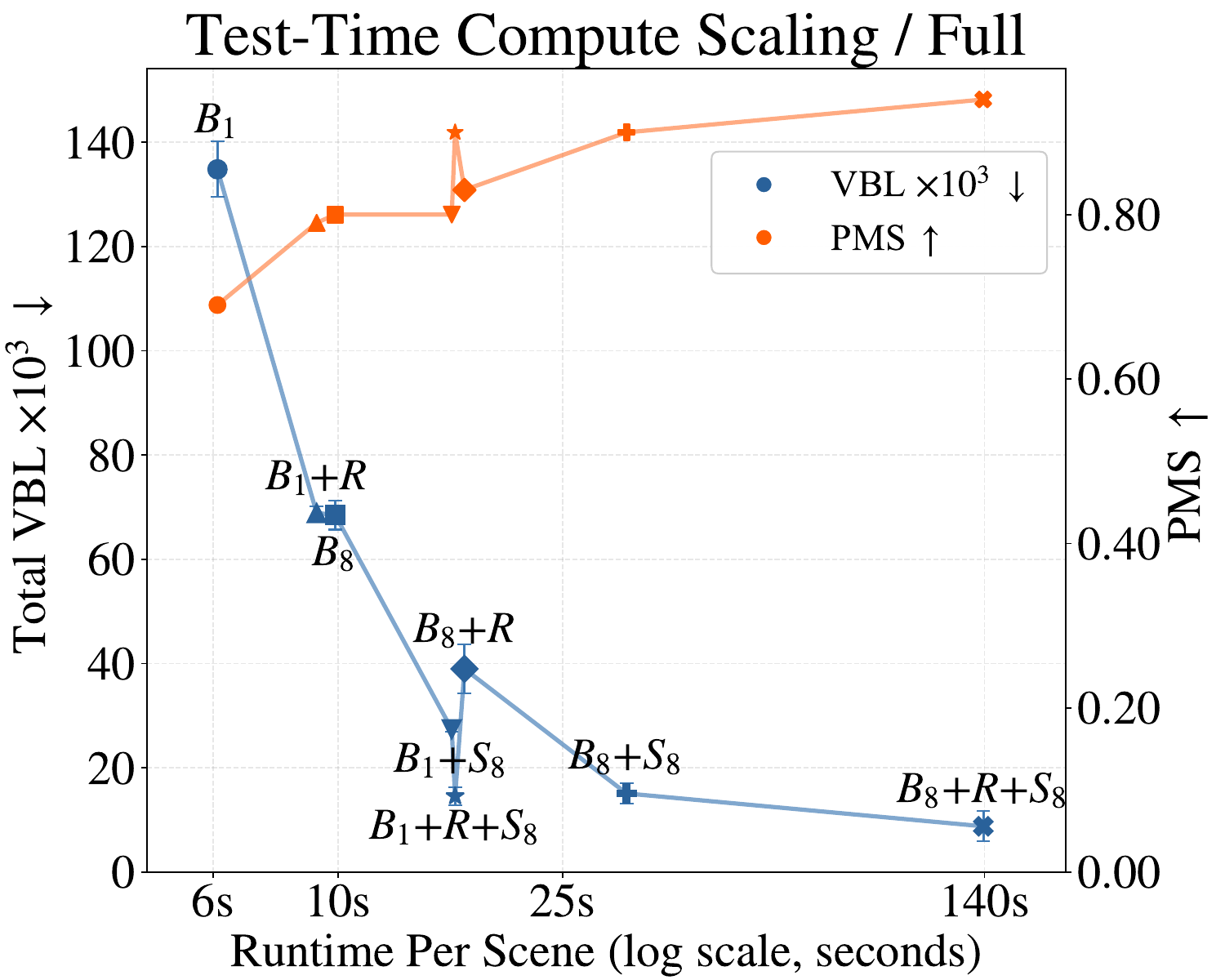} \\
    \end{tabular}
    \vspace{-2mm}
    \caption{%
        \textbf{(Top-left)} Editing accuracy on autoregressive editing ($L\leq10$).
        \textbf{(Top-right)} Delta VBL vs.\ \# of objects; with more uniform performance for ours.
        \textbf{(Bottom-left)} PMS vs.\ prompt word count, with slight negative correlation.
        \textbf{(Bottom-right)} Test-Time Scaling: impact on VBL and PMS vs.\ runtime.
    }
    \label{fig:analysis-grid}
\end{figure}

\paragraph{(i) Scene Complexity.} We study the effect of room size and existing object count on addition by clustering the number of objects per scene and aggregating them into uniform bins. We show the trend on Delta VBL for this in Fig.~\ref{fig:analysis-grid} (Top-right). As object count grows, the SSR context lengthens, increasing the long-context reasoning demands on SG-LLM. However, scenes with higher object count are also larger, with potentially more free space. We argue that an ideal model has uniform performance across varying object count, floor area, and scene density, and can see that ours performs much stronger compared to the baselines.

\vspace{-4mm}

\paragraph{(ii) Prompt Complexity.} We aggregate prompts for full scene synthesis by word count and report PMS per bin in Fig.~\ref{fig:analysis-grid} (Bottom-left). Longer prompts impose more simultaneous constraints on SG-LLM (e.g., color, material, shape, and style), and a slight but consistent negative correlation with PMS is confirmed, suggesting that instruction-following becomes marginally harder as prompt complexity increases.

\vspace{-4mm}

\paragraph{(iii) Scaling Test-Time Compute.} We explore three scaling axes on the \texttt{`bed'} split: BoN sampling ($B_8$; $N{=}8$), rotation (${+}R$; with ${\times}4$ rotated variants per addition), and shuffling ($+S_8$; with 8 random object orderings). For BoN and rotation, candidates are merged and filtered by highest PMS then lowest VBL; for shuffling, scenes are generated independently and selected by lowest VBL. As shown in Fig.~\ref{fig:analysis-grid} (Bottom-right), both $B_8$ and $B_1{+}R$ halve VBL over $B_1$ with modest PMS gains, while shuffling yields a further ${\sim}2.5{\times}$ reduction. Gains quickly saturate after, placing them on the Pareto front of quality vs. runtime. A human evaluation study (see \ref{supp:user-study-2}) confirms that $B_1{+}R{+}S_8$ achieves the strongest winrate.

\vspace{-4mm}

\paragraph{(iv) Asset-Agnostic Spatial Reasoning.} We emphasize that we use VBL primarily as \textit{evaluation} metric, for best model selection on the validation split, and as a binary reward signal during RLVR, since we cannot backpropagate through it. Unlike bounding boxes, VBL captures fine-grained geometric interactions, which matter for learning correct spatial semantics even if infrequent across the full dataset. Supported by human evaluations (\ref{supp:user-study-3}), this suggests that this choice encodes sufficient spatial constraints for effective scene arrangement, even without explicit geometric modeling, validating a key hypothesis: SG-LLM can learn effective spatial reasoning purely from structured text, enabling deployment across arbitrary asset catalogs without requiring visual input or mesh-based geometry — critical for real-world scenarios where available physical inventory may vary and asset catalogs can be swapped.

\vspace{-4mm}

\paragraph{(v) Limitations.} While ReSpace demonstrates the potential of a semantically rich scene abstraction that is independent of any specific 3D asset catalog, the current formulation reflects design choices that define the scope of this work and suggest directions for future extensions. The autoregressive placement can lead to sequences where early object placements constrain later additions, motivating exploration of more globally optimized placement. Our experiments also focus on single-room scenes and furniture, leaving extensions to multi-room and architectural elements (e.g., doors/windows). In addition, scene size is bounded by compute, not method: the released dataset has up to $50$ objects, while our $3$K-token context (given our available compute) exposes up to ${\sim}21$ objects at training. Finally, position and pose editing is currently achieved by removal followed by re-addition, with native in-place spatial edits a natural extension.


\section{Conclusion}
\label{chap:end}
We introduced \textsc{ReSpace}, a framework for text-driven 3D indoor scene synthesis and editing with autoregressive language models. Our structured scene representation encodes explicit boundaries and positioning alongside textual descriptions for objects, while our specialized SG-LLM surpasses state-of-the-art on object addition metrics. By leveraging a zero-shot LLM for object removal and prompt generation, we demonstrate superior human-perceived scene quality for full scene synthesis without end-to-end training. We further evaluate ReSpace on autoregressive editing sequences mimicking realistic user interactions, analyzing performance across varying sequence lengths and compounding errors. This paradigm opens several promising research directions: developing a single model that handles all scene synthesis and editing tasks while maintaining prompt-following capabilities; exploring scaling laws with larger context windows and model sizes as more training data becomes available; incorporating local correction via optimization after each autoregressive step to eliminate layout violations while preserving generative diversity; and investigating advanced test-time compute techniques such as Monte Carlo Tree Search to optimize full scene synthesis using PMS and VBL as verifiable rewards, moving beyond greedy addition.

{
    \small
    \bibliographystyle{ieeenat_fullname}
    \bibliography{main}
}

\section{Appendix}

This Appendix provides the following:
\begin{itemize}
    \item Details on the preprocessing of the 3D-FRONT dataset, instruction generation, and the creation of our SSR-3DFRONT dataset (Section~\ref{supp:preprocessing}),
    \item Details on preference alignment via RLVR, including GRPO, DPO, and RFT (Section~\ref{supp:rlvr}),
    \item Implementation details for training our SG-LLM (Section~\ref{supp:impl-details}),
    \item Details on stochastic asset sampling with qualitative results (Section~\ref{supp:asset-sampling}),
    \item Quantitative results on full scene synthesis (Section~\ref{supp:full-scene-quant}), 
    \item Additional qualitative samples for full scene synthesis (Section~\ref{supp:qual-samples-full}),
    \item User Study 1: human evaluation of preference alignment methods (Section~\ref{supp:user-study-1}),
    \item User Study 2: human evaluation of test-time compute scaling (Section~\ref{supp:user-study-2}),
    \item User Study 3: human evaluation of ReSpace vs.\ baselines (Section~\ref{supp:user-study-3}),
    \item Runtime analysis on full scene synthesis (Section~\ref{supp:runtime-analysis}),
    \item Removal operation analysis (Section~\ref{supp:removal-analysis}),
    \item Example of full SSR instance (Section~\ref{supp:ssr-example}), and
    \item Prompts for zero-shot LLM for command decomposition and removal (Section~\ref{supp:prompts-zero-shot}).
\end{itemize}

\subsection{Dataset Preprocessing for SSR-3DFRONT}
\label{supp:preprocessing}

Our training data curation is based on the 3D-FRONT \cite{fu20213d_front} dataset, which includes $\thicksim$19K synthetic indoor scenes of varying size and density, alongside positioned objects referenced from 3D-FUTURE \cite{fu20213d_future}, a furniture asset catalog providing textured 3D meshes and renderings for each asset. In order to bring the scenes from the 3D-FRONT dataset \cite{fu20213d_front} into our Structured Scene Representation (SSR), we proceed as follows: First, we leverage an existing dataset of postprocessed 3D room meshes from 3D-FRONT from \cite{dahnert2021panoptic} that has simplified wall geometry and closed holes, which facilitates the room boundary extraction. We run a custom line search algorithm (pseudo-code shown in Algorithm \ref{alg:rectilinear-extraction}) on each room mesh to extract an ordered set of 3D vertices forming a rectilinear polygon for the floor and ceiling. These set of vertices build the room boundaries $\mathcal{B}_\text{top}$ and $\mathcal{B}_\text{bottom}$. Next, we convert the scenes into JSON, resembling our proposed SSR (see Section \ref{chap:ssr}). We define 3 room types: \texttt{`bedroom'} for bedrooms, \texttt{`livingroom'} for living rooms, dining rooms and living/dining rooms, and \texttt{`other'} for all remaining rooms. We only consider scenes that have valid scene boundaries with $|\{b_i\}| \geq 4$, an object count of $3 \leq |\{o_i\}| \leq 50$, and $\text{VBL} < 0.1$. The latter filters out ``invalid" scenes that contain too many object or boundary collisions (see Section \ref{chap:lay-viol}). Since a few scenes have only a single malpositioned object, we further check for scene validity if a single object already violates this filter (via $\text{VBL} \geq 0.1$), and keep the modified scene if removing that object makes the scene valid. We shift all scenes to the origin $(0,0,0)$. In total, this results in $13055$ valid scenes after preprocessing. Our dataset is available \href{https://huggingface.co/datasets/gradient-spaces/SSR-3DFRONT}{here}.



\begin{algorithm}
\caption{Rectilinear Polygon Corner Extraction}
\label{alg:rectilinear-extraction}
\begin{algorithmic}[1]

\Require Mesh vertices $V \in \mathbb{R}^{n \times 3}$
\Ensure Corner vertices $C$ forming rectilinear polygon

\State $P \leftarrow \text{unique}(V[:, [0,1]])$ \Comment{Project to unique 2D points}
\State $(x_0, y_0) \leftarrow \argmin_{v: v_x = \min(P[:, 0])} v_y$, $\text{dir} \leftarrow \text{'north'}$, $\text{curr} \leftarrow (x_0, y_0)$, $C \leftarrow []$

\Repeat
    \State $S \leftarrow \text{getSortedAxisPoints}(P, \text{curr}, \text{dir})$ \Comment{Points on same axis, sorted by direction}
    \State $i \leftarrow \text{indexOf}(\text{curr}, S) + 1$ 
    \While{$\neg\text{isCornerPoint}(P, S[i], \text{dir})$} $i \leftarrow i + 1$ \EndWhile
    \State $\text{curr} \leftarrow S[i]$, $\text{dir} \leftarrow \text{getNextDirection}(P, \text{curr}, \text{dir})$, $C.\text{append}(\text{curr})$
\Until{curr = $(x_0, y_0)$}
\State \Return $C$

\end{algorithmic}
\end{algorithm}

\subsubsection{Prompts for Asset Description and Prompt Bank}
\label{supp:prompts-assets}
As mentioned in Section \ref{chap:experiments}, the raw 3D-FRONT dataset does not contain textual descriptions of assets or scenes, and we leverage GPT-4o \cite{hurst2024gpt} as a Vision-Language Model (VLM) to generate sentence-level descriptions $\bf{d}_j$ for each object $\bf{o}_j$ in the catalog. We query the VLM by attaching a rendering of the asset with a prompt that includes the provided class label. After obtaining sentence-level descriptions for each generation, we leverage the same VLM to generate 10 unique, concise prompts (2-5 words in noun phrase format) for each asset description $d_j$. This approach serves two purposes: (i) it covers diverse prompting styles with varying levels of detail and word order, and (ii) it prevents trivial overfitting on our small dataset by avoiding repetition of identical prompts that lead to memorization rather than generalization. We provide the full prompt for extracting visual properties in textual form for each asset in the 3D-FUTURE dataset \cite{fu20213d_future} in Figure \ref{fig:asset-desc-prompt} and for prompt generation in Figure\ref{fig:asset-prompt-bank}. For asset descriptions, we leverage the content provided in the \texttt{`summary'} as it seemed to best capture dense semantics that refer to style, color, material, etc.

\begin{figure}
    \begin{tcolorbox}[top=2pt,bottom=2pt, width=\linewidth, boxrule=1pt, halign=left]
        {\scriptsize {\fontfamily{zi4}\selectfont
        \textbf{User Prompt} \\
Please provide a concise JSON object of the furniture item in the image using `style', `color', `material', `characteristics', and `summary' as keys. Describe the style, noting any blends of design elements. Specify the materials used for different components (if applicable). List the key characteristics, including the shape, design features, and any distinctive elements or decorative accents. If there are multiple values for a key, use a list of strings. DO NOT build a nested JSON. The summary compactly captures the essence of the furniture’s style, functionality, and aesthetic appeal, emphasizing its unique attributes. This description should clearly differentiate this piece from others while succinctly capturing its essential properties and we will use it for object retrieval, so it should be as accurate as possible, keyword-heavy, but just be one extremely short sentence. You are an interior designer EXPERT. Hint: It's a \{ASSET\_OBJECT\_CATEGORY\_LABEL\}. Only output the JSON as a plain string and nothing else.
        }
        \par}
    \end{tcolorbox}
    \vspace{-2mm}
    \caption{Prompt for GPT-4o \cite{hurst2024gpt} for extracting various object properties for each asset including a sentence-level asset description, given a rendering of an object in the 3D-FUTURE dataset \cite{fu20213d_future}.}
    \label{fig:asset-desc-prompt}
\end{figure}

\begin{figure}
    \begin{tcolorbox}[top=2pt,bottom=2pt, width=\linewidth, boxrule=1pt, halign=left]
        {\scriptsize {\fontfamily{zi4}\selectfont
        \textbf{User Prompt} \\
The list below contains a sentence referring to a single piece of furniture. Your task it to create a list of 10 short descriptions that vary in length. Each description refers to the subject with a maximum of 3-4 additional descriptive words that reference the color, style, shape, etc. All your sentences should be in `noun phrase'. You MUST include a variety of lengths in your descriptions, ensuring a few samples are very short (1-2 words max) and others are longer (4-5 words). Have at least one sample with only one word, except if you need to be more specific for the subject, e.g., use `Coffee Table', not just `Table', if present. Use mostly basic properties such as color or material, but also include a few creative and diverse versions to increase robustness in our ML training dataset.\\

The sentence is:\\

- \{ASSET\_DESCRIPTION\} \\

Just output a plain list and nothing else. You have only one list of 10 descriptions. You MUST always point to the referenced object above and not hallucinate other furniture or be overly generic by using `furniture' or `piece'. Every list contains the descriptions in increasing word length. Just output the final JSON object as a plain string without any key. Never use markdown or \textasciigrave \textasciigrave \textasciigrave json.
        }
        \par}
    \end{tcolorbox}
    \vspace{-2mm}
    \caption{Prompt for GPT-4o \cite{hurst2024gpt} for generating a prompt bank with a list of 10 unique prompts, given a sentence-level asset description.}
    \label{fig:asset-prompt-bank}
\end{figure}

\subsubsection{Instruction Generation for SG-LLM}
\label{supp:instr-gen}
Given the full scenes, we impose dynamic instruction generations based on a stochastic recipe. Let $\mathcal P(o)=\{p_1,\dots,p_K\}$ be the fixed prompt bank for object $o$ (we set $K=10$). During training, we turn $\mathcal S$ into an instruction tuple: $\mathcal I=\bigl(\hat{\mathcal S},\;p,\;o_{\text{add}}\bigr)$, where the model must learn to add object $o_{\text{add}}$ to the partial scene $\hat{\mathcal S} = \bigl(\mathcal T,\mathcal B,\hat{\mathcal O}\bigr)$ when conditioned on the natural-language prompt $p$. To generate a tuple, we first draw a random permutation for the order of objects $\pi\thicksim\mathrm{Unif}(S_N)$, then uniformly sample the prompt $p\thicksim\mathrm{Unif}\!\bigl(\mathcal P(o_{\text{add}})\bigr)$ for object $o_{\text{add}}$. Let $Z \in \{Z_0, Z_1, Z_2\}$ be the instruction type: $Z_0$ (\texttt{`zero\_start'}), $Z_1$ (\texttt{`full\_scene'}), and $Z_2$ (\texttt{`random'}), with $Z_0$ teaching the model to start from an empty room given only the prompt, $Z_1$ teaching `scene completion' as the final contains all objects from the scene but $o_{\text{add}}$, and $Z_2$ teaching robust object placements on arbitrary, shuffled partial scenes. For $Z_0$ we set $\hat{\mathcal O}=\varnothing,\;o_{\text{add}}=o_{\pi(1)}$. For $Z_1$ we set $\hat{\mathcal O}=\{o_{\pi(1)},\dots,o_{\pi(N-1)}\},\;o_{\text{add}}=o_{\pi(N)}$. For $Z_2$ we draw a drop count $M\thicksim\mathrm{Unif}\{0,\dots,N-1\}$, put $L=N-M$ and define $o_{\text{add}}=o_{\pi(L)},\;\hat{\mathcal O}=\{o_{\pi(1)},\dots,o_{\pi(L-1)}\}$. Instruction type is sampled as $Z\thicksim\mathrm{Cat}(w_0,w_1,w_2)$ with fixed $w_0=w_1=0.1,\;w_2=0.8$ and the conditional distribution factorizes as

\vspace{-3mm}

\[
  p(\mathcal I\mid\mathcal S)
  =\sum_{z=0}^{2} w_z\,p(\mathcal I\mid Z=z,\mathcal S),
\]


\[
  p(\mathcal{I} \mid Z=z, \mathcal{S}) =
  \begin{cases}
    \displaystyle\frac{\mathbf{1}_{\{z=0\}}}{N\,|\mathcal{P}(o_{\text{add}})|} & (z=0) \\[10pt]
    \displaystyle\frac{\mathbf{1}_{\{z=1\}}}{N!\,|\mathcal{P}(o_{\text{add}})|} & (z=1) \\[10pt]
    \displaystyle\frac{\mathbf{1}_{\{z=2\}}}{N!\,N\,|\mathcal{P}(o_{\text{add}})|} & (z=2)
  \end{cases}
\]

\vspace{2mm}

Thus, for $Z_2$, we choose one of $N!$ permutations, one of $N$ drop counts, and one of $|\mathcal P(o_{\text{add}})|$ prompts. Fixed weights $w_0=w_1=0.1$ guarantee at least 20\% exposure to the empty-room and near-complete-room edge cases even for very large scenes. Since we have have empty or full scenes with $\frac{1}{N}$ probability (and partial scenes otherwise), their likelihood decreases inversely proportional with higher object count. Imposing minimum exposure via fixed weights ensures the model learns these edge cases as well. We perform random data augmentation on train/val samples by (i) rotating each scene by $\theta \in \{0, 90, 180, 270 \}^\circ$, (ii) cyclically shifting room bounds in a round-robin fashion, and (iii) slightly perturbing $x-$ and $z-$components of every position and size vector of each object with a uniform delta with $v' = v + \delta$ and $\delta \thicksim \mathrm{U}(-0.02, 0.02)$ for coordinate values $v$.

\subsection{Preference Alignment via RLVR}
\label{supp:rlvr}
After training via SFT, we experiment with preference alignment on our existing base model. SG-LLM is trained via SFT on single object addition with synthetic triples (\textit{scene}, \textit{prompt}, \textit{obj\_add}), where the existing scene is passed as SSR together with an object-level prompt as input, and the output \textit{obj\_add} represents the next placed object. Given the object-level prompt, the placed object, the current scene, and the known ground-truth object, we can deterministically assign \textit{verifiable rewards} for a single object placement --- enabling preference alignment directly on our SFT model without human intervention. We define high-quality placement as:

\begin{equation}
\label{eq:reward_filter}
\mathcal{F}(a_i) = \mathbf{1}\!\left[
\begin{array}{l}
    \text{PMS}(a_i) \geq 0.85 \\[4pt]
    \wedge\; \text{VBL}(a_i) < 10^{-5} \\[4pt]
    \wedge\; \left\|\dfrac{s_{a_i} - s_{a_i}^{\text{GT}}}{s_{a_i}^{\text{GT}}}\right\|_2 < 0.5
\end{array}
\right],
\end{equation}

\vspace{2mm}

\noindent where $\text{PMS}(a_i)$ measures prompt-following quality defined as Prompt Matching Score (see Section \ref{chap:exp-settings}), $\text{VBL}(a_i)$ quantifies layout violations via our voxelization-based loss (see section \ref{chap:lay-viol}), and $\left\|(s_{a_i} - s_{a_i}^{\text{GT}}) / s_{a_i}^{\text{GT}}\right\|_2$ is the relative L2 distance between the predicted and ground-truth 3D bounding box size, normalizing each dimension by its ground-truth value. This binary filter $\mathcal{F}$ serves as our reward signal: placements satisfying all criteria receive a reward of $+1$, and all others $0$.

Without this strict filter, we observed strong reward hacking: the model produced structurally valid JSONs but learned to generated 3d bounding box size such that smaller assets got sampled, which trivially reduces intersection probability and drives VBL toward zero, while degrading prompt adherence. Imposing our joint filter, particularly the size L2 constraint, closes this loophole by penalizing size deviations from ground truth regardless of collision behavior. However, this choice is not perfect, since a good model ideally captures a strong multi-modal distribution with differently sized objects as valid placements (possibly resulting in dist $\geq 0.5$).

\paragraph{\textbf{(i) GRPO}}
(Group Relative Policy Optimization). Introduced in \cite{shao2024deepseekmath}, let for each iteration be $G$ candidates $a_i$ with verifiable reward $r_i$ and the objective:

\vspace{-4mm}

\begin{equation}
\begin{aligned}
&r_i(\theta) = \tfrac{\pi_\theta(a_i|s)}{\pi_{\text{old}}(a_i|s)}, \quad
 \hat{r}_i = \text{clip}(r_i, 1{-}\varepsilon, 1{+}\varepsilon) \\
&J(\theta) = \frac{1}{G}\sum_{i=1}^G
  \Big(\min(r_i A_i,\, \hat{r}_i A_i) - \beta\, D_{\text{KL}}(\pi_\theta \| \pi_{r})\Big)
\end{aligned}
\end{equation}


where each term in the sum is expanded as a per-token loss per response $a_i$. The advantage $A_i$ is given as $A_i = \frac{r_i - \text{mean}({r_1,\cdots, r_G})}{\text{std}({r_1,\cdots, r_G})}$, $\beta$ controls the KL divergence between the current policy $\pi_\theta$ and reference policy $\pi_{r}$, and $\varepsilon$ is given as upper/lower-bound for clipping.

For GRPO fine-tuning, we use an LR of 5e-5, $\text{temp}=0.7$, batch size of 6, GAS of 16, 6 generations per sample/instruction, and set $\beta=0.0$ to cancel out the KL divergence term. We select the best model based on lowest delta VBL after already 3 epochs. We give rewards of $-1.0$ for invalid JSON outputs and $1.0$ for candidates $a_i$ that pass our quality filter. One of the main advantages of GRPO is that it can leverage more than 2 rollouts per input, as it does not rely on pair-wise comparisons, and we use 6 rollouts per input/output pair. Since high-quality samples only appear with around $25\%$ probability, negative rewards dominate and corrupt the SFT behavior too aggressively — especially the JSON structure. Thus, we employ a \textit{high-quality-only distillation} that cancels out the loss for samples with valid JSON but that do not satisfy our filter. 

\paragraph{\textbf{(ii) DPO}}
(Direct Preference Optimization). Originally introduced by \cite{rafailov2023direct} as an offline alternative to RLHF \cite{ouyang2022training}, DPO performs pairwise comparisons of two outputs for the same input, eliminating the need for a separate reward model. Rather than assigning scalar advantages per response, DPO designates a chosen response $r_A$ and a rejected response $r_B$, and directly optimizes the policy to increase the relative likelihood of $r_A$ over $r_B$:

\begin{equation}
\begin{aligned}
\mathcal{L}_{\text{DPO}}(\theta) = -\mathbb{E}_{(s,r_A,r_B)}\!\left[\log \sigma\!\left(\right.\right.
  &\beta \log \tfrac{\pi_\theta(r_A|s)}{\pi_{\text{ref}}(r_A|s)} \\
  - &\beta \log \tfrac{\pi_\theta(r_B|s)}{\pi_{\text{ref}}(r_B|s)}
\left.\left.\right)\right]
\end{aligned}
\end{equation}

where $\sigma$ is the sigmoid function, $\beta$ controls the deviation from the reference policy $\pi_{\text{ref}}$, and $(s, r_A, r_B)$ are triples of input scene, chosen, and rejected object placements. We generate two candidates per input and assign chosen/rejected using filter $\mathcal{F}$ (Eq.~\ref{eq:reward_filter}). Analogously to our GRPO setup, when both candidates pass or both fail $\mathcal{F}$ — providing no clear preference signal — we mask out the loss for that pair rather than introducing noise via random assignment. We use a LR of 2e-5, batch size of 2, GAS of 32, $\beta=0.1$, and 512 random samples from the training set per epoch. We select the best checkpoint based on lowest delta VBL, reached after 87 epochs.

\paragraph{\textbf{(iii) RFT}}
(Rejection Sampling Fine-Tuning). Rather than optimizing a policy objective directly, RFT iteratively generates $N$ candidate completions per training sample, filters them using $\mathcal{F}$ (Eq.~\ref{eq:reward_filter}), and fine-tunes the model via SFT on the accepted samples. For each round, we use a fixed subset of 512 scenes, rather than re-sampling randomly each epoch, so that the model can observe quality improvements on the same inputs over successive rounds — which we found empirically to outperform random sampling for each epoch. For samples where no candidate passes $\mathcal{F}$, we fall back to the ground-truth completion to prevent the model from catastrophically forgetting difficult placements. To avoid overrepresenting easy samples that consistently produce many accepted completions, we cap accepted samples per prompt at $K=2$ and deduplicate before capping. We use an LR of 1e-5, batch size of 4, GAS of 16, $N=16$ generations per prompt, resulting in the best checkpoint with lowest delta VBL after 27 epochs.

\subsection{Implementation Details for SG-LLM}
\label{supp:impl-details}
For baselines, we use the released source code, modify the pre-processing to fit our custom dataset splits, and re-train Mi-Diff and ATISS from scratch on our three different datasets. We do not modify their hyperparameter choice and pick the best model based on their lowest validation loss.

We trained SG-LLM on a two-stage pipeline via SFT+RLVR. For the first stage, we perform Supervised Fine-Tuning (SFT) on the full weights for 30-50 hours with a learning rate (LR) of 5e-5, local batch size of 4, gradient accumulation step (GAS) of 8, and a context window of $3000$ tokens, selecting the model with best validation loss via mean delta VBL on the val split. We experimented with Low-Rank Adaptation (LoRA) \cite{hu2022lora} but observed faster convergence with SFT on full weights. We used 4xA100 NVIDIA 80GB GPUs with 16 CPUs and 384GB RAM, running Python 3.9.0 with CUDA 12.1.1, GCC 10.3.0, and \texttt{`bf16'} numerical precision. We conducted extensive experiments with 0.5B/1B/1.5B models, and observed best results with \texttt{`Qwen2.5-1.5B-Instruct'} \cite{yang2024qwen2}, together with \texttt{`Llama-3.1-8B-Instruct'} \cite{grattafiori2024llama} for the zero-shot LLM. Using vLLM \cite{kwon2023efficient} during inference speeds up generation on full scenes especially beyond $BoN{=}1$.

\subsection{Stochastic Asset Sampling}
\label{supp:asset-sampling}
\begin{figure}
    \centering
    \includegraphics[width=1.0\linewidth]{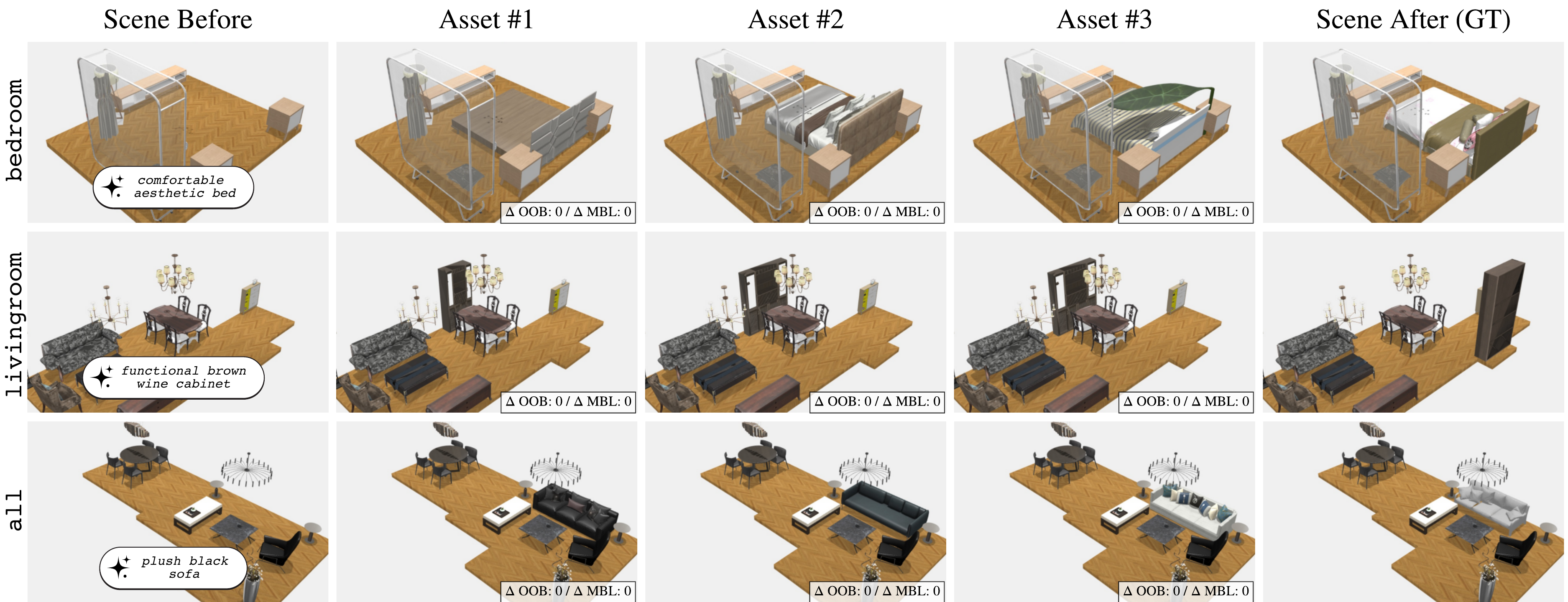}
    \vspace{-5mm}
    \caption{Qualitative results for stochastic asset sampling without greedy selection (as done for the samples shown in Figure \ref{fig:instr-qualitative-samples}).}
    \label{fig:full-qualitative-samples-assets}
\end{figure}

Our proposed stochastic asset sampling involves various hyperparameters to tweak the final discrete distribution. We heuristically found that $\lambda=0.5$, $\sigma=0.2$, $\text{temp}=0.2$, $\text{top\_p}=0.95$, and $\text{top\_k=20}$ perform the best. However, for all experiments reported in the results of the main paper (see Section \ref{chap:experiments}), we impose a \textit{greedy selection} strategy for asset sampling in order to maintain better comparison with baselines, consistent with LLM evaluation practices that use low-temperature decoding for reproducible comparisons without sampling variance \cite{bucher2024fine}. We use the same hyperparameters as above and choose the top-1 asset via $\text{argmax}_{m_j}g_{\phi}(d_i, h_i)$. Additionally, we show true asset sampling (with the same hyperparameters) in Figure \ref{fig:full-qualitative-samples-assets} on the same instructions from Figure \ref{fig:instr-qualitative-samples} and 3 randomly sampled assets. We can simply sample from the distribution (instead of top-1 selection) for true stochasticity. We suggest that there might not be a single set of best hyperparameters for asset sampling. Instead, the user might tweak $\lambda$, (i.e., the strength of the semantic embedding), or $\sigma$ (i.e., the sharpness of the size matching via 3D bounding boxes) dynamically during scene generation to guide the process towards more desired candidates. With $\lambda=0.5$, both geometry (via 3d bounding box size differences) and semantics (via SigLIP embeddings) have equal contribution to the final distribution for samples picked in Figure \ref{fig:full-qualitative-samples-assets}.

\subsection{Quantitative Results on Full Scenes}
\label{supp:full-scene-quant}
As discussed in Section~\ref{chap:full-scene-synth}, we additionally evaluate ReSpace on full scene synthesis against end-to-end trained baselines. Table ~\ref{exps-scenes-full} report results for all room types on the full test set split (non-rectangular rooms as well).

\begin{table*}
\caption{Quantitative evaluation on \textbf{full scenes} using 500 unseen floor plans with 3 random seeds per sample. Metrics follow Table \ref{exps-scenes-instr}.}
  \vspace{-2mm}
  \label{exps-scenes-full}
  \centering
  \scriptsize
  \resizebox{\textwidth}{!}{%
  \begin{tabular}{@{}c l*{7}{r}@{}}
    \toprule
    & & \multicolumn{3}{c}{Layout Violations} & \multicolumn{3}{c}{Scene Renderings} & \multicolumn{1}{c}{Prompt} \\
    \cmidrule(r){3-5} \cmidrule(r){6-8} \cmidrule(r){9-9}
    & Method & \multicolumn{1}{c}{$\text{OOB}_{\times\text{1e3}} \downarrow$} & \multicolumn{1}{c}{$\text{MBL}_{\times\text{1e3}} \downarrow$} & \multicolumn{1}{c}{$\text{VBL}_{\times\text{1e3}} \downarrow$} & \multicolumn{1}{c}{$\text{FID} \downarrow$} & \multicolumn{1}{c}{$\text{FID}_\text{CLIP} \downarrow$} & \multicolumn{1}{c}{$\text{KID}_{\times\text{1e3}} \downarrow$} & \multicolumn{1}{c}{$\text{PMS} \uparrow$} \\
    
    \midrule
    
    \multirow{4}{*}{\rotatebox[origin=c]{90}{\texttt{`bed'}}} 
    
    & ATISS 
    & $414.3_{\pm 23.6}$ 
    & $99.7_{\pm 6.4}$ 
    & $514.1_{\pm 24.4}$ 
    & $\underline{43.51}_{\pm .3}$ 
    & $\underline{2.34}_{\pm .1}$ 
    & $2.51_{\pm .5}$ 
    & $n/a$ \\
    
    & Mi-Diff 
    & $360.1_{\pm 18.0}$ 
    & $\underline{66.6}_{\pm 9.5}$ 
    & $427.0_{\pm 08.5}$ 
    & $\mathbf{43.18}_{\pm .4}$ 
    & $\mathbf{2.23}_{\pm .1}$ 
    & $\mathbf{1.34}_{\pm .2}$ 
    & $n/a$ \\
    
    & $\text{ReSpace/A}^{\dagger}$ 
    & $\underline{62.8}_{\pm 02.7}$ 
    & $72.0_{\pm 5.1}$ 
    & $\underline{134.8}_{\pm 05.3}$ 
    & $45.33_{\pm 1.}$ 
    & $2.79_{\pm .1}$ 
    & $2.91_{\pm .5}$ 
    & $\underline{0.69}_{\pm .0}$ \\

    & $\text{ReSpace/A}^{\dagger}_{S8{+}R}$
    & $\mathbf{2.9}_{\pm 01.0}$ 
    & $\mathbf{11.7}_{\pm 1.3}$ 
    & $\mathbf{14.6}_{\pm 01.7}$ 
    & $44.21_{\pm .2}$ 
    & $2.72_{\pm .1}$ 
    & $\underline{2.62}_{\pm .3}$
    & $\mathbf{0.90}_{\pm .0}$ \\

    \midrule
    
    \multirow{4}{*}{\rotatebox[origin=c]{90}{\texttt{`liv'}}} 
    
    & ATISS 
    & $506.6_{\pm 22.2}$ 
    & $135.1_{\pm 6.1}$ 
    & $641.6_{\pm 28.2}$ 
    & $44.14_{\pm .3}$ 
    & $\underline{2.26}_{\pm .0}$ 
    & $8.06_{\pm .3}$ 
    & $n/a$ \\
    
    & Mi-Diff 
    & $361.5_{\pm 12.7}$ 
    & $\underline{117.1}_{\pm 3.2}$ 
    & $478.7_{\pm 09.6}$ 
    & ${40.76}_{\pm .1}$ 
    & $\mathbf{2.11}_{\pm .1}$ 
    & ${4.29}_{\pm .1}$ 
    & $n/a$ \\
    
    & $\text{ReSpace/A}^{\dagger}$ 
    & $\underline{158.4}_{\pm 03.0}$ 
    & $159.7_{\pm 11.}$ 
    & $\underline{318.1}_{\pm 10.8}$ 
    & $\underline{40.75}_{\pm .1}$ 
    & $2.54_{\pm .1}$ 
    & $\underline{3.18}_{\pm .3}$ 
    & $\underline{0.70}_{\pm .0}$ \\

    & $\text{ReSpace/A}^{\dagger}_{S8{+}R}$
    & $\mathbf{4.5}_{\pm 00.6}$ 
    & $\mathbf{22.7}_{\pm 2.7}$ 
    & $\mathbf{27.2}_{\pm 03.1}$ 
    & $\mathbf{39.73}_{\pm .3}$ 
    & $3.30_{\pm .1}$ 
    & $\mathbf{2.78}_{\pm .1}$
    & $\mathbf{0.86}_{\pm .0}$ \\
    
    \midrule

    \multirow{4}{*}{\rotatebox[origin=c]{90}{\texttt{`all'}}} 
    
    & ATISS 
    & $631.4_{\pm 12.9}$ 
    & $108.5_{\pm 6.9}$ 
    & $739.8_{\pm 19.0}$ 
    & $45.58_{\pm .1}$ 
    & $\underline{2.37}_{\pm .0}$ 
    & $3.87_{\pm .1}$ 
    & $n/a$ \\
    
    & Mi-Diff 
    & $327.4_{\pm 41.3}$ 
    & $\underline{87.1}_{\pm 2.7}$ 
    & $414.5_{\pm 41.6}$ 
    & $\mathbf{42.57}_{\pm .3}$ 
    & $\mathbf{2.14}_{\pm .0}$ 
    & $\underline{1.27}_{\pm .2}$ 
    & $n/a$ \\
    
    & $\text{ReSpace/A}^{\dagger}$ 
    & $\underline{92.8}_{\pm 12.4}$ 
    & $98.1_{\pm 8.2}$ 
    & $\underline{190.9}_{\pm 20.6}$ 
    & ${43.15}_{\pm .2}$ 
    & $2.46_{\pm .1}$ 
    & $\mathbf{1.26}_{\pm .1}$ 
    & $\underline{0.71}_{\pm .0}$ \\

    & $\text{ReSpace/A}^{\dagger}_{S8{+}R}$
    & $\mathbf{4.2}_{\pm 00.8}$ 
    & $\mathbf{14.6}_{\pm 3.9}$ 
    & $\mathbf{18.7}_{\pm 04.7}$ 
    & $\underline{42.96}_{\pm .5}$ 
    & $2.64_{\pm .1}$ 
    & ${1.58}_{\pm .1}$
    & $\mathbf{0.90}_{\pm .0}$ \\
    
    \bottomrule
  \end{tabular}%
  }
\vspace{-3mm}
\end{table*}

\subsection{More Qualitative Examples on Full Scenes}
\label{supp:qual-samples-full}
\begin{figure}
    \centering
    \includegraphics[width=1.0\linewidth]{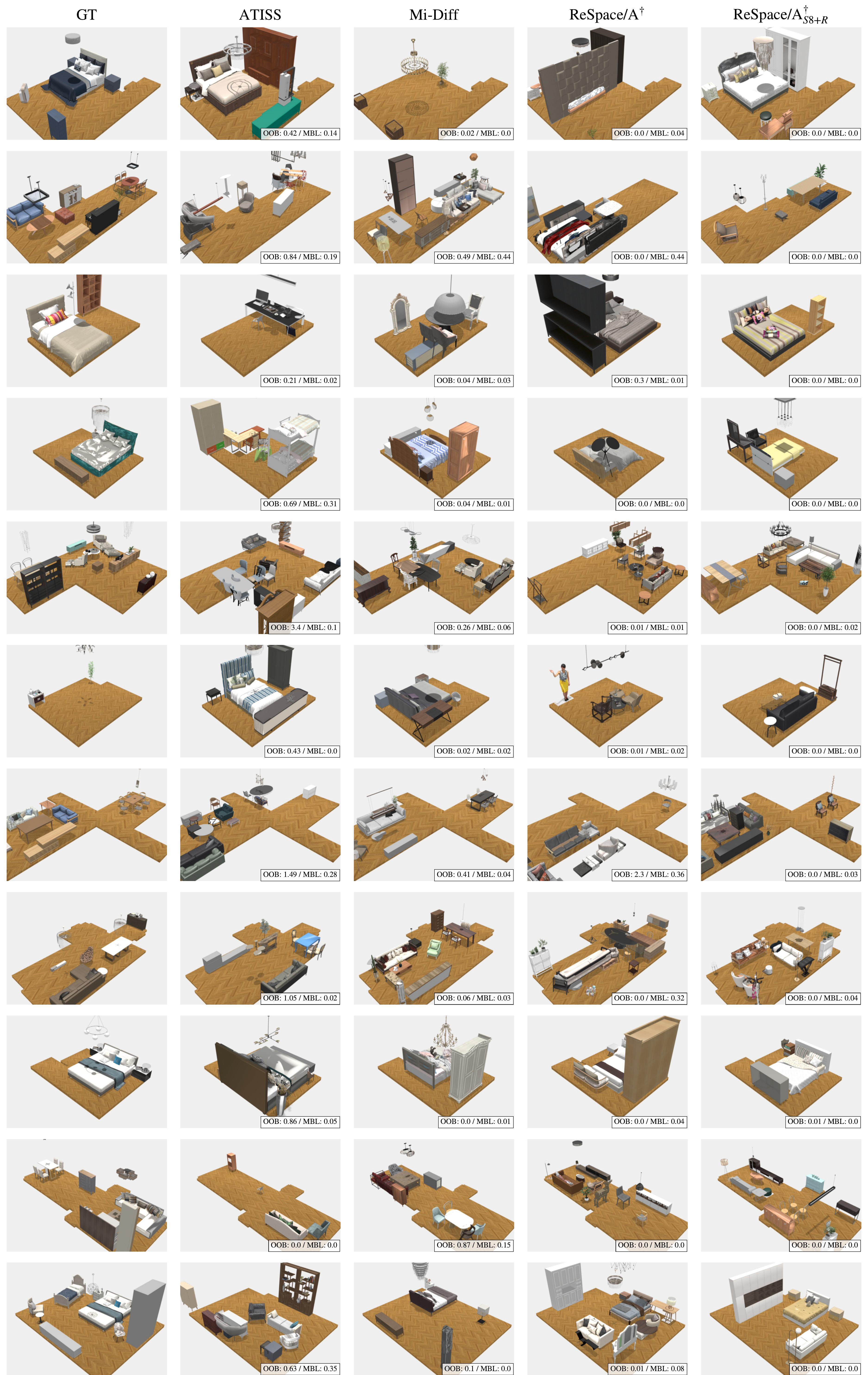}
    \caption{Qualitative results (random selection) on full scene synthesis with baselines.}
    \label{fig:full-qualitative-samples-supp}
\end{figure}

In Figure \ref{fig:full-qualitative-samples-supp}, we show additional qualitative samples for full scene synthesis (with greedy asset sampling; otherwise same setup as done in the main experiments in Section \ref{chap:experiments}). In contrast to Figure \ref{fig:full-qualitative-samples}, we show both the vanilla version of our method (no test-time compute scaling; BoN=1, no shuffling) and with shuffling enabled. Due to  randomness in our full pipeline between different runs (i.e., especially involving the zero-shot LLM for the prompt list generation and the sampling of few shot samples and number of objects via priors), results for ours do not involve the same object prompt lists and result in different scene-level compositions. Despite this fact, the results with shuffling enabled show better scene quality compared to BoN=1, with less OOB and MBL, and better overall composition.

\subsection{User Study 1: Preference Alignment}
\label{supp:user-study-1}

To evaluate the impact of various preference alignment strategies for SG-LLM on human-perceived scene quality, we conducted a targeted human evaluation study comparing our SFT baseline against various post-training variants: GRPO, DPO, and RFT (Rejection Sampling Fine-Tuning). Table~\ref{tab:user_study_ablation_pairwise} summarizes the results.

\begin{table}
\centering
\caption{User Study 1: Human evaluation results via Bradley-Terry analysis for ablation on various preference alignment methods.}
\vspace{-2mm}
\footnotesize
\label{tab:user_study_ablation_pairwise}
\begin{tabular}{cccc}
\toprule
Method A & Method B & Win Rate A & Win Rate B \\
\midrule
SFT & GRPO & 49.0\% & {51.0\%} \\
SFT & RFT & 41.2\% & \textbf{58.8\%} \\
SFT & DPO & 47.9\% & {52.1\%} \\
\bottomrule
\end{tabular}
\end{table}

We conduct a user study to evaluate the effect of preference alignment on human-perceived scene quality, running it in two phases. In the first phase, 150 participants performed 4,500 pairwise comparisons between SFT and GRPO, finding no statistically significant preference (51\% vs.\ 49\%). This confirms that while GRPO provides directional improvements in layout violations as measured by VBL, it does not translate to a meaningful change in human-perceived scene quality. This is consistent with the training fragility we observed: too aggressive preference optimization risks corrupting JSON generation and spatial reasoning, while too conservative an update does not meaningfully modify SFT behavior. In the second phase, 226 participants performed 6,780 pairwise comparisons across three method pairs: SFT vs.\ RFT, SFT vs.\ DPO, and RFT vs.\ DPO. RFT consistently outperforms both SFT and DPO, with a modest but consistent margin over SFT (58.8\% vs.\ 41.2\%) and a win over DPO (55.9\% vs.\ 44.1\%). DPO shows only a marginal preference over SFT (52.1\% vs.\ 47.9\%), suggesting it provides limited improvements over the SFT baseline despite achieving a significantly lower VBL for both single object placements and full scene synthesis. Taken together, these results establish RFT as the most effective preference alignment strategy for improving human-perceived scene quality over the SFT baseline. We hypothesize that although DPO achieved the strongest filter pass rate (i.e., the highest fraction of object additions satisfying our high-quality placement threshold), optimizing directly against preference pairs may overfit to the reward signal for single object placement without improving the broader scene-coherent placement that human evaluators respond to. We show a full quantitative comparison between all our models (following the same setting as in \ref{chap:experiments}) on single object addition in Table \ref{exps:scenes-instr-extensive-ours} and full scene synthesis in Table \ref{exps:scenes-full-extensive-ours}.

\begin{table*}
  \caption{Full quantitative evals on \textbf{single object addition} (same as in Table \ref{exps-scenes-instr}).}
  \vspace{-2mm}
  \centering
  \scriptsize
  \label{exps:scenes-instr-extensive-ours}
  \resizebox{\textwidth}{!}{%
  \begin{tabular}{@{}c l*{7}{r}@{}}
    \toprule
    & & \multicolumn{3}{c}{Layout Violations} & \multicolumn{3}{c}{Scene Renderings} & \multicolumn{1}{c}{Prompting} \\
    \cmidrule(r){3-5} \cmidrule(r){6-8} \cmidrule(r){9-9}
    & Method & \multicolumn{1}{c}{$\text{OOB}^{\Delta}_{\times1e3} \downarrow$} & \multicolumn{1}{c}{$\text{MBL}^{\Delta}_{\times1e3} \downarrow$} & \multicolumn{1}{c}{$\text{VBL}^{\Delta}_{\times1e3} \downarrow$} & \multicolumn{1}{c}{$\text{FID} \downarrow$} & \multicolumn{1}{c}{$\text{FID}_\text{CLIP} \downarrow$} & \multicolumn{1}{c}{$\text{KID}_{\times\text{1e3}} \downarrow$} & \multicolumn{1}{c}{$\text{PMS} \uparrow$} \\
    
    \midrule
    
    \multirow{4}{*}{\rotatebox[origin=c]{90}{\texttt{`bed'}}} 
    
    & ATISS & $97.70_{\pm 6.0}$ & $13.54_{\pm 0.5}$ & $111.24_{\pm 5.4}$ & $36.18_{\pm .3}$ & $1.74_{\pm .0}$ & $0.19_{\pm .0}$ & $0.58_{\pm .0}$ \\
    
    & Mi-Diff & $64.04_{\pm 5.3}$ & $14.27_{\pm 1.5}$ & $78.31_{\pm 4.1}$ & $36.12_{\pm .3}$ & $1.76_{\pm .1}$ & ${0.05}_{\pm .0}$ & $0.57_{\pm .0}$ \\

    \hdashline
    


    & $\text{ReSpace/B}_{\text{SFT}}$
    & ${11.77}_{\pm 3.7}$ 
    & ${4.45}_{\pm 0.5}$ 
    & ${16.23}_{\pm 4.0}$ 
    & ${\underline{35.23}}_{\pm .3}$ 
    & ${\textbf{1.64}}_{\pm .0}$ 
    & ${\underline{-0.06}}_{\pm .0}$ 
    & ${\underline{0.88}}_{\pm .0}$ \\

    & $\text{ReSpace/A}_{\text{SFT}}$ 
    & ${13.89}_{\pm 0.3}$ 
    & ${\underline{3.77}}_{\pm 0.7}$ 
    & ${17.66}_{\pm 0.6}$ 
    & ${\textbf{35.22}}_{\pm .1}$ 
    & ${1.67}_{\pm .0}$ 
    & ${\textbf{-0.13}}_{\pm .0}$ 
    & ${\underline{0.88}}_{\pm .0}$ \\
    
    & $\text{ReSpace/A}_{\text{GRPO}}$ 
    & ${\underline{8.38}}_{\pm 1.5}$ 
    & ${8.13}_{\pm 1.2}$ 
    & ${16.51}_{\pm 3.0}$ 
    & ${35.51}_{\pm .2}$ 
    & ${1.67}_{\pm .0}$ 
    & ${0.06}_{\pm .1}$ 
    & ${\textbf{0.89}}_{\pm .0}$ \\


    & $\text{ReSpace/A}_{\text{RFT}}$ 
    & ${10.75}_{\pm 2.6}$ 
    & ${3.91}_{\pm 0.7}$ 
    & ${\underline{14.66}}_{\pm 2.4}$ 
    & ${35.35}_{\pm .2}$ 
    & ${\underline{1.66}}_{\pm .0}$ 
    & ${-0.03}_{\pm .1}$ 
    & ${\textbf{0.89}}_{\pm .0}$ \\


    & $\text{ReSpace/A}_{\text{DPO}}$
    & ${\textbf{4.19}}_{\pm 2.1}$ 
    & ${\textbf{2.82}}_{\pm 0.5}$ 
    & ${\textbf{7.00}}_{\pm 2.5}$ 
    & ${35.70}_{\pm .0}$ 
    & ${1.70}_{\pm .0}$ 
    & ${0.12}_{\pm .0}$ 
    & ${0.83}_{\pm .2}$ \\
    
    \midrule
    
    \multirow{4}{*}{\rotatebox[origin=c]{90}{\texttt{`liv'}}} 
    
    & ATISS & $63.87_{\pm 6.9}$ & $11.43_{\pm 3.8}$ & $75.30_{\pm 5.8}$ & $32.26_{\pm .1}$ & $1.48_{\pm .0}$ & ${0.71}_{\pm .3}$ & $0.58_{\pm .0}$ \\
    
    & Mi-Diff & $43.88_{\pm 7.6}$ & $12.87_{\pm 1.4}$ & $56.75_{\pm 8.8}$ & $33.30_{\pm .3}$ & $1.53_{\pm .0}$ & $1.06_{\pm .2}$ & $0.56_{\pm .0}$ \\

    \hdashline
    


    & $\text{ReSpace/L}_{\text{SFT}}$
    & ${10.68}_{\pm 0.5}$ 
    & ${4.27}_{\pm 0.3}$ 
    & ${14.95}_{\pm 0.7}$ 
    & ${31.94}_{\pm .0}$ 
    & ${\underline{1.40}}_{\pm .0}$ 
    & ${\textbf{0.21}}_{\pm .1}$ 
    & ${\underline{0.84}}_{\pm .0}$ \\

    & $\text{ReSpace/A}_{\text{SFT}}$
    & ${12.30}_{\pm 1.6}$ 
    & ${4.99}_{\pm 0.4}$ 
    & ${17.29}_{\pm 1.5}$ 
    & ${\underline{31.88}}_{\pm .1}$ 
    & ${\textbf{1.39}}_{\pm .0}$ 
    & ${\underline{0.22}}_{\pm .1}$ 
    & ${0.85}_{\pm .0}$ \\
    
    & $\text{ReSpace/A}_{\text{GRPO}}$ 
    & ${11.20}_{\pm 3.1}$ 
    & ${8.22}_{\pm 1.0}$ 
    & ${19.41}_{\pm 4.1}$ 
    & ${31.90}_{\pm .2}$ 
    & ${\underline{1.40}}_{\pm .0}$ 
    & ${0.19}_{\pm .2}$ 
    & ${0.87}_{\pm .0}$ \\


    
    & $\text{ReSpace/A}_{\text{RFT}}$
    & ${\textbf{7.51}}_{\pm 1.7}$ 
    & ${\underline{3.11}}_{\pm 0.9}$ 
    & ${\textbf{10.62}}_{\pm 2.5}$
    & ${\textbf{31.84}}_{\pm .0}$ 
    & ${1.41}_{\pm .0}$ 
    & ${1.41}_{\pm .0}$ 
    & ${0.87}_{\pm .0}$ \\
    

    & $\text{ReSpace/A}_{\text{DPO}}$
    & ${\underline{8.77}}_{\pm 3.1}$ 
    & ${\textbf{2.84}}_{\pm 0.9}$ 
    & ${\underline{11.61}}_{\pm 3.6}$ 
    & ${31.94}_{\pm .0}$ 
    & ${1.42}_{\pm .0}$ 
    & ${0.32}_{\pm .2}$ 
    & ${\textbf{0.79}}_{\pm .0}$ \\
    
    \midrule
    
    \multirow{3}{*}{\rotatebox[origin=c]{90}{\texttt{`all'}}} 
    
    & ATISS & $121.66_{\pm 8.6}$ & ${14.48}_{\pm 1.0}$ & $136.14_{\pm 8.7}$ & $36.40_{\pm .0}$ & $1.77_{\pm .0}$ & $0.22_{\pm .1}$ & ${0.57}_{\pm .0}$ \\
    
    & Mi-Diff & ${40.51}_{\pm 5.5}$ & $18.19_{\pm 0.6}$ & ${58.70}_{\pm 4.9}$ & ${36.14}_{\pm .2}$ & ${1.72}_{\pm .0}$ & ${0.07}_{\pm .1}$ & $0.56_{\pm .0}$ \\

    \hdashline


    & $\text{ReSpace/A}_{\text{SFT}}$ 
    & ${17.37}_{\pm 4.4}$ 
    & ${5.09}_{\pm 1.0}$ 
    & ${22.47}_{\pm 4.4}$ 
    & ${\underline{35.45}}_{\pm .1}$ 
    & ${\textbf{1.66}}_{\pm .0}$ 
    & ${\textbf{-0.14}}_{\pm .1}$ 
    & ${\underline{0.86}}_{\pm .0}$ \\
    
    & $\text{ReSpace/A}_{\text{GRPO}}$ 
    & ${13.11}_{\pm 3.7}$ 
    & ${8.67}_{\pm 2.3}$ 
    & ${21.78}_{\pm 6.0}$ 
    & ${35.71}_{\pm .4}$ 
    & ${\underline{1.67}}_{\pm .0}$ 
    & ${-0.03}_{\pm .1}$ 
    & ${\textbf{0.87}}_{\pm .0}$ \\


    
    & $\text{ReSpace/A}_{\text{RFT}}$
    & ${\underline{7.61}}_{\pm 1.8}$ 
    & ${\underline{3.60}}_{\pm 1.0}$ 
    & ${\underline{11.21}}_{\pm 2.3}$ 
    & ${\textbf{35.41}}_{\pm .3}$ 
    & ${\textbf{1.66}}_{\pm .0}$ 
    & ${\underline{-0.06}}_{\pm .1}$ 
    & ${\textbf{0.87}}_{\pm .0}$ \\
    

    & $\text{ReSpace/A}_{\text{DPO}}$
    & ${\textbf{3.87}}_{\pm 1.2}$ 
    & ${\textbf{2.39}}_{\pm 1.0}$ 
    & ${\textbf{6.26}}_{\pm 2.1}$ 
    & ${35.66}_{\pm .2}$ 
    & ${1.71}_{\pm .0}$ 
    & ${0.21}_{\pm .0}$ 
    & ${0.80}_{\pm .0}$ \\
    
    \bottomrule
  \end{tabular}%
    }
\end{table*}

\begin{table*}
\caption{Full quantitative evaluations on \textbf{full scenes} (same as in Table \ref{exps-scenes-full}).}
  \vspace{-2mm}
  \scriptsize
  \label{exps:scenes-full-extensive-ours}
  \resizebox{\textwidth}{!}{%
  \begin{tabular}{@{}c l*{7}{r}@{}}
    \toprule
    & & \multicolumn{3}{c}{Layout Violations} & \multicolumn{3}{c}{Scene Renderings} & \multicolumn{1}{c}{Prompt} \\
    \cmidrule(r){3-5} \cmidrule(r){6-8} \cmidrule(r){9-9}
    & Method & \multicolumn{1}{c}{$\text{OOB}_{\times\text{1e3}} \downarrow$} & \multicolumn{1}{c}{$\text{MBL}_{\times\text{1e3}} \downarrow$} & \multicolumn{1}{c}{$\text{VBL}_{\times\text{1e3}} \downarrow$} & \multicolumn{1}{c}{$\text{FID} \downarrow$} & \multicolumn{1}{c}{$\text{FID}_\text{CLIP} \downarrow$} & \multicolumn{1}{c}{$\text{KID}_{\times\text{1e3}} \downarrow$} & \multicolumn{1}{c}{$\text{PMS} \uparrow$} \\
    
    \midrule
    
    \multirow{3}{*}{\rotatebox[origin=c]{90}{\texttt{`bed'}}} 
    & ATISS 
    & $414.3_{\pm 23.6}$ 
    & $99.7_{\pm 6.4}$ 
    & $514.1_{\pm 24.4}$ 
    & ${43.51}_{\pm .3}$ 
    & ${\underline{2.34}}_{\pm .1}$ 
    & $2.51_{\pm .5}$ 
    & $n/a$ \\
    
    & Mi-Diff 
    & $360.1_{\pm 18.0}$ 
    & ${\underline{66.6}}_{\pm 9.5}$ 
    & $427.0_{\pm 08.5}$ 
    & ${\textbf{43.18}}_{\pm .4}$ 
    & ${\textbf{2.23}}_{\pm .1}$ 
    & ${\textbf{1.34}}_{\pm .2}$ 
    & $n/a$ \\

    \hdashline


    & $\text{ReSpace/A}_{\text{SFT}}$
    & ${93.8}_{\pm 01.7}$ 
    & ${71.3}_{\pm 12.}$ 
    & ${165.1}_{\pm 13.0}$ 
    & $\underline{43.50}_{\pm .4}$ 
    & $2.55_{\pm .0}$ 
    & $\underline{2.04}_{\pm .2}$ 
    & ${\underline{0.69}}_{\pm .0}$ \\
    
    & $\text{ReSpace/A}_{\text{GRPO}}$ 
    & ${67.4}_{\pm 07.1}$ 
    & $140.7_{\pm 20.}$ 
    & ${208.1}_{\pm 13.4}$ 
    & $44.77_{\pm .2}$ 
    & $2.70_{\pm .0}$ 
    & ${2.17}_{\pm .1}$ 
    & ${\textbf{0.72}}_{\pm .0}$ \\



    & $\text{ReSpace/A}_{\text{RFT}}$ 
    & ${\underline{62.8}}_{\pm 02.7}$ 
    & ${72.0}_{\pm 5.1}$ 
    & ${\underline{134.8}}_{\pm 05.3}$ 
    & $45.33_{\pm 1.}$ 
    & $2.79_{\pm .1}$ 
    & $2.91_{\pm .5}$ 
    & ${\underline{0.69}}_{\pm .0}$ \\


    & $\text{ReSpace/A}_{\text{DPO}}$
    & ${\textbf{49.8}}_{\pm 7.82}$ 
    & ${\textbf{52.6}}_{\pm 1.7}$ 
    & ${\textbf{102.4}}_{\pm 7.89}$ 
    & $46.31_{\pm .6}$ 
    & $3.02_{\pm .0}$ 
    & $4.05_{\pm .1}$ 
    & ${0.65}_{\pm .0}$ \\
    
    \midrule
    
    \multirow{3}{*}{\rotatebox[origin=c]{90}{\texttt{`liv'}}} 
    & ATISS 
    & $506.6_{\pm 22.2}$ 
    & $135.1_{\pm 6.1}$ 
    & $641.6_{\pm 28.2}$ 
    & $44.14_{\pm .3}$ 
    & ${\underline{2.26}}_{\pm .0}$ 
    & $8.06_{\pm .3}$ 
    & $n/a$ \\
    
    & Mi-Diff 
    & $361.5_{\pm 12.7}$ 
    & ${\textbf{117.1}}_{\pm 3.2}$ 
    & ${478.7}_{\pm 09.6}$ 
    & ${\underline{40.76}}_{\pm .1}$ 
    & ${\textbf{2.11}}_{\pm .1}$ 
    & ${\underline{4.29}}_{\pm .1}$ 
    & $n/a$ \\

    \hdashline

    & $\text{ReSpace/A}_{\text{SFT}}$
    & ${243.9}_{\pm 10.7}$ 
    & $202.4_{\pm 1.8}$ 
    & $446.3_{\pm 10.6}$ 
    & $44.05_{\pm .5}$ 
    & $2.50_{\pm .1}$ 
    & $7.22_{\pm .6}$ 
    & ${0.66}_{\pm .2}$ \\
    
    & $\text{ReSpace/A}_{\text{GRPO}}$ 
    & ${254.4}_{\pm 14.2}$ 
    & $310.4_{\pm 13.}$ 
    & $564.8_{\pm 25.8}$ 
    & $46.17_{\pm .3}$ 
    & $2.42_{\pm .1}$ 
    & $8.05_{\pm .7}$ 
    & ${\textbf{0.73}}_{\pm .0}$ \\



    & $\text{ReSpace/A}_{\text{RFT}}$ 
    & ${\underline{158.4}}_{\pm 03.0}$ 
    & $159.7_{\pm 11.}$ 
    & $\underline{318.1}_{\pm 10.8}$ 
    & $\textbf{40.75}_{\pm .1}$ 
    & $2.54_{\pm .1}$ 
    & $\textbf{3.18}_{\pm .3}$ 
    & ${\underline{0.70}}_{\pm .0}$ \\


    & $\text{ReSpace/A}_{\text{DPO}}$
    & ${\textbf{148.8}}_{\pm 10.8}$ 
    & $\underline{122.9}_{\pm 7.4}$ 
    & $\textbf{270.8}_{\pm 15.5}$ 
    & $42.07_{\pm .5}$ 
    & $2.83_{\pm .1}$ 
    & $4.70_{\pm .2}$ 
    & ${0.62}_{\pm .0}$ \\
    
    \midrule
    
    \multirow{3}{*}{\rotatebox[origin=c]{90}{\texttt{`all'}}} 
    & ATISS 
    & $631.4_{\pm 12.9}$ 
    & $108.5_{\pm 6.9}$ 
    & $739.8_{\pm 19.0}$ 
    & $45.58_{\pm .1}$ 
    & ${2.37}_{\pm .0}$ 
    & $3.87_{\pm .1}$ 
    & $n/a$ \\
    
    & Mi-Diff 
    & $327.4_{\pm 41.3}$ 
    & ${\underline{87.1}}_{\pm 2.7}$ 
    & $414.5_{\pm 41.6}$ 
    & ${\textbf{42.57}}_{\pm .3}$ 
    & ${\textbf{2.14}}_{\pm .0}$ 
    & ${\underline{1.27}}_{\pm .2}$ 
    & $n/a$ \\

    \hdashline


    & $\text{ReSpace/A}_{\text{SFT}}$
    & ${125.4}_{\pm 09.9}$ 
    & ${116.6}_{\pm 12.}$ 
    & ${241.9}_{\pm 22.0}$ 
    & $43.53_{\pm .4}$ 
    & $\underline{2.33}_{\pm .0}$ 
    & $1.94_{\pm .4}$ 
    & ${\underline{0.68}}_{\pm .0}$ \\
    
    & $\text{ReSpace/A}_{\text{GRPO}}$ 
    & ${160.2}_{\pm 16.0}$ 
    & $181.6_{\pm 26.}$ 
    & ${341.8}_{\pm 17.9}$ 
    & $44.85_{\pm .2}$ 
    & $2.43_{\pm .2}$ 
    & $2.44_{\pm .5}$ 
    & ${\textbf{0.71}}_{\pm .0}$ \\



    & $\text{ReSpace/A}_{\text{RFT}}$
    & ${\underline{92.8}}_{\pm 12.4}$ 
    & ${98.1}_{\pm 8.2}$ 
    & ${\underline{190.9}}_{\pm 20.6}$ 
    & ${\underline{43.15}}_{\pm .2}$ 
    & ${2.46}_{\pm .1}$ 
    & ${\textbf{1.26}}_{\pm .1}$ 
    & ${\textbf{0.71}}_{\pm .0}$ \\


    & $\text{ReSpace/A}_{\text{DPO}}$
    & ${\textbf{74.7}}_{\pm 04.5}$ 
    & ${\textbf{77.4}}_{\pm 9.7}$ 
    & ${\textbf{152.2}}_{\pm 09.8}$ 
    & ${44.49}_{\pm .6}$ 
    & ${2.59}_{\pm .1}$ 
    & ${2.57}_{\pm .3}$ 
    & ${0.62}_{\pm .0}$ \\
    
    \bottomrule
  \end{tabular}%
  }
\end{table*}

\subsection{User Study 2: Test-Time Scaling}
\label{supp:user-study-2}
We conduct a second human evaluation study to assess the effect of test-time compute scaling on human-perceived scene quality for our method. As described in Section~\ref{chap:experiments}, we explore three scaling axes: BoN sampling ($B_8$), rotation (${+}R$), and shuffling (${+}S_8$). Quantitative results on the \texttt{`bed'} split across all eight configurations are reported in Table~\ref{exps-scenes-full-ablation-ttc-ours}, showing consistent VBL reductions along each axis. The study involved 3 method variants evaluated on the \texttt{`bed'} split: $B_1$ (no scaling), $B_1{+}S_8$ (shuffling only), and $B_1{+}R{+}S_8$ (rotation and shuffling). Participants were shown pairs of generated scenes and asked to select which appeared more coherent, with Bradley-Terry analysis used to rank methods.

\begin{table}[h]
\centering
\caption{User Study 2: Human evaluation results via Bradley-Terry analysis for test-time compute scaling on full scene synthesis (\texttt{`bed'} split).}
\vspace{-2mm}
\footnotesize
\label{tab:user_study_ttc}
\setlength{\tabcolsep}{6pt}
\begin{tabular}{cllll}
\toprule
Rank & Method & BT Score & Std Dev & Win Rate \\
\midrule
1 & ReSpace ($B_1{+}R{+}S_8$) & 0.4143 & 0.0115 & \textbf{58.7\%} \\
2 & ReSpace ($B_1{+}S_8$)     & 0.3175 & 0.0099 & 48.9\% \\
3 & ReSpace ($B_1$)           & 0.2682 & 0.0087 & 42.6\% \\
\bottomrule
\end{tabular}
\end{table}

The study involved 2,986 individual pairwise comparisons. $B_1{+}R{+}S_8$ achieves the strongest win rate at 58.7\%, outperforming both $B_1{+}S_8$ (48.9\%) and $B_1$ (42.6\%) in direct pairwise comparisons. Concretely, $B_1{+}R{+}S_8$ wins 616 out of 994 comparisons against $B_1$ ($62.0\%$) and 541 out of 978 against $B_1{+}S_8$ ($55.3\%$), while $B_1{+}S_8$ wins 537 out of 1,014 comparisons against $B_1$ ($53.0\%$). These results confirm that both shuffling and rotation contribute meaningfully to human-perceived scene quality, consistent with the VBL reductions observed in Table~\ref{exps-scenes-full-ablation-ttc-ours}. Notably, $B_1{+}R{+}S_8$ achieves this at only ${\sim}16$s per scene, placing it firmly on the Pareto front of quality versus runtime (see Section \ref{supp:runtime-analysis}).

\begin{table*}
\caption{Full quantitative evaluations on full scenes and various test-time compute scaling axes (\texttt{'bed'} split)}.
  \vspace{-2mm}
  \scriptsize
  \label{exps-scenes-full-ablation-ttc-ours}
  \resizebox{\textwidth}{!}{%
  \begin{tabular}{@{}c l*{7}{r}@{}}
    \toprule
    & & \multicolumn{3}{c}{Layout Violations} & \multicolumn{3}{c}{Scene Renderings} & \multicolumn{1}{c}{Prompt} \\
    \cmidrule(r){3-5} \cmidrule(r){6-8} \cmidrule(r){9-9}
    & Method & \multicolumn{1}{c}{$\text{OOB}_{\times\text{1e3}} \downarrow$} & \multicolumn{1}{c}{$\text{MBL}_{\times\text{1e3}} \downarrow$} & \multicolumn{1}{c}{$\text{VBL}_{\times\text{1e3}} \downarrow$} & \multicolumn{1}{c}{$\text{FID} \downarrow$} & \multicolumn{1}{c}{$\text{FID}_\text{CLIP} \downarrow$} & \multicolumn{1}{c}{$\text{KID}_{\times\text{1e3}} \downarrow$} & \multicolumn{1}{c}{$\text{PMS} \uparrow$} \\
    
    \midrule
    
    \multirow{4}{*}{\rotatebox[origin=c]{90}{\texttt{`bed'}}} 
    
    & ATISS 
    & $414.3_{\pm 23.6}$ 
    & $99.7_{\pm 6.4}$ 
    & $514.1_{\pm 24.4}$ 
    & ${43.51}_{\pm .3}$ 
    & ${2.34}_{\pm .1}$ 
    & $2.51_{\pm .5}$ 
    & $n/a$ \\
    
    & Mi-Diff 
    & $360.1_{\pm 18.0}$ 
    & ${66.6}_{\pm 9.5}$ 
    & $427.0_{\pm 08.5}$ 
    & ${43.18}_{\pm .4}$ 
    & ${2.23}_{\pm .1}$ 
    & ${1.34}_{\pm .2}$ 
    & $n/a$ \\

    \hdashline
    
    & $\text{ReSpace/A}^{\dagger} ({B1 \cdot - \cdot -})$ 
    & ${62.8}_{\pm 02.7}$ 
    & $72.0_{\pm 5.1}$ 
    & ${134.8}_{\pm 05.3}$ 
    & $45.33_{\pm 1.}$ 
    & $2.79_{\pm .1}$ 
    & $2.91_{\pm .5}$ 
    & ${0.69}_{\pm .0}$ \\
    
    & $\text{ReSpace/A}^{\dagger} ({B1 \cdot R \cdot -})$ 
    & ${23.2}_{\pm 01.9}$ 
    & $45.7_{\pm 0.8}$ 
    & ${68.9}_{\pm 01.3}$ 
    & $44.98_{\pm .7}$ 
    & $2.72_{\pm .0}$ 
    & $2.08_{\pm .3}$ 
    & ${0.79}_{\pm .0}$ \\

    \hdashline
    
    & $\text{ReSpace/A}^{\dagger} ({B1 \cdot - \cdot S8})$ 
    & ${7.9}_{\pm 00.6}$ 
    & ${19.5}_{\pm 0.7}$ 
    & ${27.4}_{\pm 00.5}$ 
    & $43.73_{\pm .5}$ 
    & $2.51_{\pm .0}$ 
    & ${2.26}_{\pm .3}$ 
    & ${0.80}_{\pm .0}$ \\

    & $\text{ReSpace/A}^{\dagger} ({B1 \cdot R \cdot S8})$ 
    & ${2.9}_{\pm 01.0}$ 
    & ${11.7}_{\pm 1.3}$ 
    & ${14.6}_{\pm 01.7}$ 
    & $44.21_{\pm .2}$ 
    & $2.72_{\pm .1}$ 
    & ${2.62}_{\pm .3}$
    & ${0.90}_{\pm .0}$ \\

    \hdashline

    & $\text{ReSpace/A}^{\dagger} ({B8 \cdot - \cdot -})$
    & ${24.7}_{\pm 03.1}$ 
    & ${43.8}_{\pm 1.2}$ 
    & ${68.5}_{\pm 02.8}$ 
    & $44.70_{\pm .8}$ 
    & $2.73_{\pm .4}$ 
    & ${2.79}_{\pm .4}$
    & ${0.80}_{\pm .2}$ \\

    & $\text{ReSpace/A}^{\dagger} ({B8 \cdot R \cdot -})$
    & ${11.7}_{\pm 01.9}$ 
    & ${27.3}_{\pm 3.9}$ 
    & ${39.0}_{\pm 04.7}$ 
    & $45.14_{\pm .5}$ 
    & $2.82_{\pm .0}$ 
    & ${3.22}_{\pm .2}$
    & ${0.83}_{\pm .2}$ \\

    \hdashline

    & $\text{ReSpace/A}^{\dagger} ({B8 \cdot - \cdot S8})$
    & ${3.4}_{\pm 01.4}$ 
    & ${11.7}_{\pm 0.7}$ 
    & ${15.1}_{\pm 02.0}$ 
    & $43.60_{\pm .2}$ 
    & $2.54_{\pm .1}$ 
    & ${2.58}_{\pm .0}$
    & ${0.90}_{\pm .0}$ \\

    & $\text{ReSpace/A}^{\dagger} ({B8 \cdot R \cdot S8})$ 
    & ${0.9}_{\pm 00.4}$ 
    & ${8.0}_{\pm 2.6}$ 
    & ${8.8}_{\pm 02.9}$ 
    & $44.21_{\pm .4}$ 
    & $2.74_{\pm .1}$ 
    & ${2.88}_{\pm .1}$
    & ${0.94}_{\pm .0}$ \\
    
    \bottomrule
  \end{tabular}%
  }
\end{table*}

\subsection{User Study 3: ReSpace vs. Baselines}
\label{supp:user-study-3}
We conduct a third human evaluation study on a rectangular-only subset of the \texttt{`all'} split to enable comparison with all baselines. The study involved 334 participants performing 10,307 pairwise comparisons across 100 randomly sampled scenes generated via full scene synthesis, using $\text{ReSpace/A}^{\dagger}$ with $B_1{+}R{+}S_8$. Participants were shown pairs of generated scenes and asked to select which appeared more coherent, with Bradley-Terry analysis used to rank methods.

\begin{figure}
    \centering
    \includegraphics[width=\linewidth]{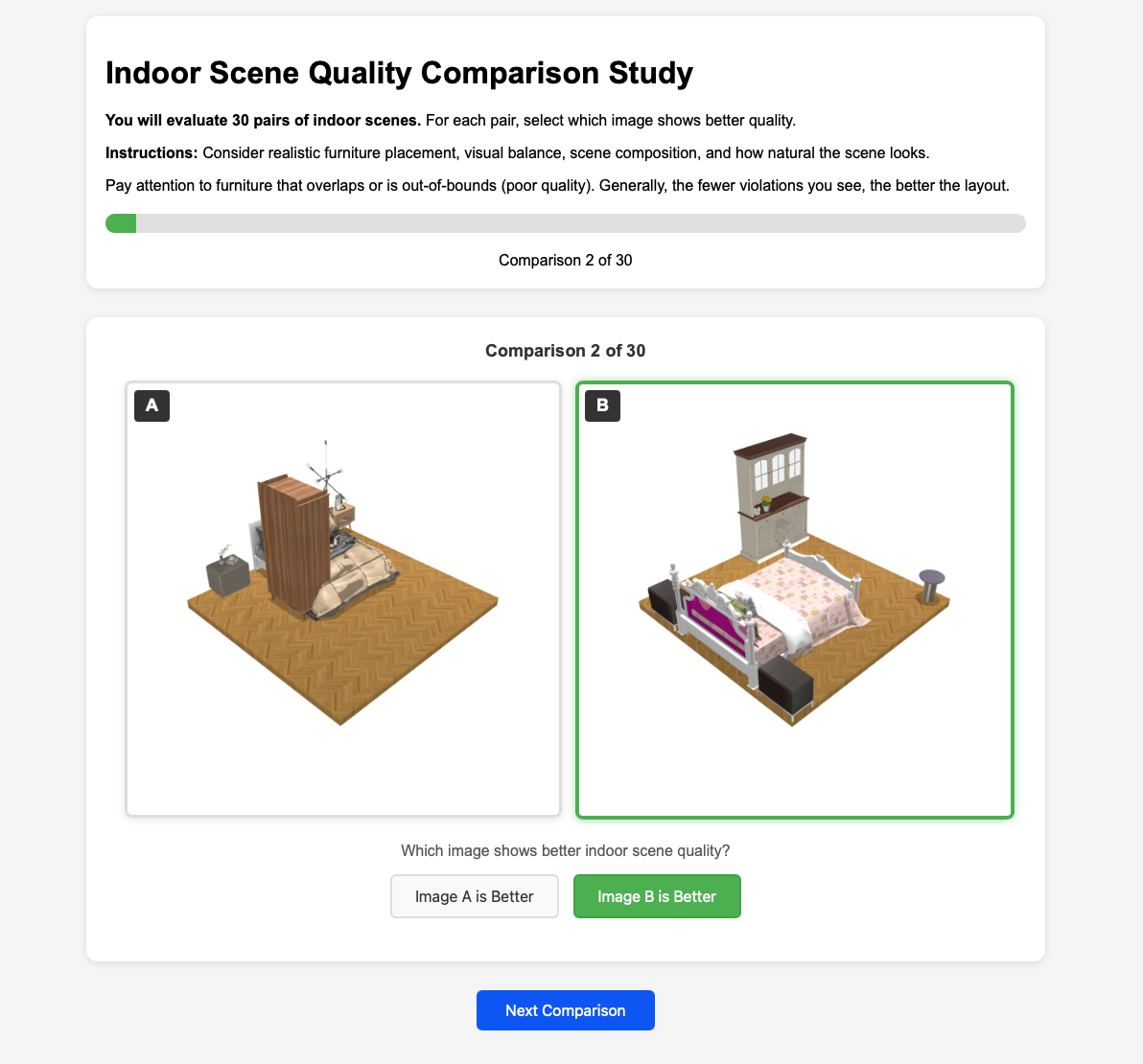}
    \vspace{-5mm}
    \caption{Interface for our pairwise human evaluations. Participants view two scenes generated from the same instruction and select the more coherent layout, with responses aggregated via Bradley-Terry.}
    \label{fig:user-study-interface}
\end{figure}

Figure~\ref{fig:user-study-interface} shows the study interface presented to participants for all user studies: a forced-choice pairwise comparison where two rendered scenes for the same instruction and room boundaries are shown side by side, without exposing the method name (i.e., fully anonymous A/B choice). The participant must then select the more coherent one. Each comparison displays the rooms from a fixed viewpoint with identical asset rendering, isolating layout quality from asset appearance.

\begin{table}[h]
\centering
\caption{User Study 3: Human evaluation results using Bradley-Terry analysis on full scene synthesis (rectangular-only subset). Results based on 10,307 pairwise comparisons from 334 participants across 100 randomly sampled scenes.}
\vspace{-2mm}
\footnotesize
\label{tab:user_study}
\begin{tabular}{r p{3cm} llll}
\toprule
Rank & Method & BT Score & Std Dev & Win Rate \\
\midrule
1 & ReSpace (ours) & 0.4251 & 0.0089 & \textbf{75.3\%} \\
2 & Mi-Diff        & 0.2143 & 0.0057 & 56.7\% \\
3 & ATISS          & 0.1486 & 0.0045 & 45.7\% \\
4 & LayoutVLM      & 0.1230 & 0.0039 & 39.9\% \\
5 & LayoutGPT      & 0.0890 & 0.0029 & 31.2\% \\
\bottomrule
\end{tabular}
\end{table}

ReSpace substantially outperforms all baselines, achieving a win rate of 75.3\% --- more than 18 percentage points above Mi-Diff (56.7\%), with direct pairwise win rates of 72.2\% against ATISS and 84.2\% against LayoutGPT. While ReSpace achieves the lowest layout violation metrics among all methods, it trades off slightly on FID/KID compared to end-to-end trained baselines --- an expected consequence of not training directly on full scene synthesis (see results in Tables \ref{exps-scenes-full} and \ref{exps-scenes-full-rect-only}). The strong human preference results confirm that this tradeoff is favorable, with further discussion in Section~\ref{chap:full-scene-synth}.

\subsection{Runtime Analysis on Full Scene Synthesis}
\label{supp:runtime-analysis}
We compare the latency of our method with other baselines to better understand the design trade-offs. We run full scene synthesis with $N{=}50$ for each method on rectangular rooms from the \texttt{`bedroom'} test set and report mean and variance in seconds in Table~\ref{tab:runtime-analysis}. ReSpace ($\text{BoN}{=}1$) achieves competitive runtime of 6.11s, outperforming LayoutGPT (6.92s) while supporting non-rectangular layouts and text-driven editing --- capabilities absent in faster feed-forward methods like ATISS and Mi-Diff. Even with $\text{BoN}{=}8$ test-time scaling, ReSpace (8.34s) remains significantly faster than LayoutVLM (32.75s). With prompt list shuffling and rotation augmentation ($S8{+}R$, $\text{BoN}{=}1$), runtime increases to 16.12s, still within half the latency of LayoutVLM. Single object addition averages 1.03s per object with $\text{BoN}{=}1$, 1.51s with $\text{BoN}{=}8$, and 3.52s with $S8{+}R$ on a single RTX 4090 GPU with 24GB. Further speedups are possible with improved vLLM inference and a larger KV cache, quantization, distillation, and further optimized VBL computation on multi-core systems.

\begin{table}
    \centering
    \caption{Runtime Analysis ours vs. baselines on full scenes}
    \vspace{-2mm}
    \footnotesize
    \label{tab:runtime-analysis}
    \footnotesize
    \setlength{\tabcolsep}{2pt}
    \begin{tabular}{@{}c p{3.5cm} r@{}}
    \toprule
    Rank & Method & Runtime (s) \\
    \midrule
    
    1 & ATISS & $00.52_{\pm00.16}$ \\
    2 & Mi-Diff & $03.70_{\pm00.19}$ \\

    3 & $\text{ReSpace/A}^{\dagger}$ ($\text{BoN}{=}1)$ & $06.11_{\pm02.80}$ \\
    
    4 & LayoutGPT & $06.92_{\pm02.56}$ \\
    
    5 & $\text{ReSpace/A}^{\dagger}$ ($\text{BoN}{=}8)$ & $08.34_{\pm04.97}$ \\
    
    6 & $\text{ReSpace/A}^{\dagger}_{S8{+R}}$ & $16.12_{\pm11.41}$ \\ 


    
    7 & LayoutVLM & $32.75_{\pm07.91}$ \\
    
    \bottomrule
    \end{tabular}
\end{table}

\subsection{Removal Operation Analysis}
\label{supp:removal-analysis}
To investigate the relatively low removal accuracy of $75.2\% \pm 1.0$ on the \texttt{`liv'} dataset compared to $90.9\% \pm 0.6$ on \texttt{`bed'} and $87.3\% \pm 0.7$ on \texttt{`all'}, we conduct a detailed analysis across three dimensions: SSR length, prompt length, and failure modes.

\begin{figure}
    \centering
    \includegraphics[width=1.0\linewidth]{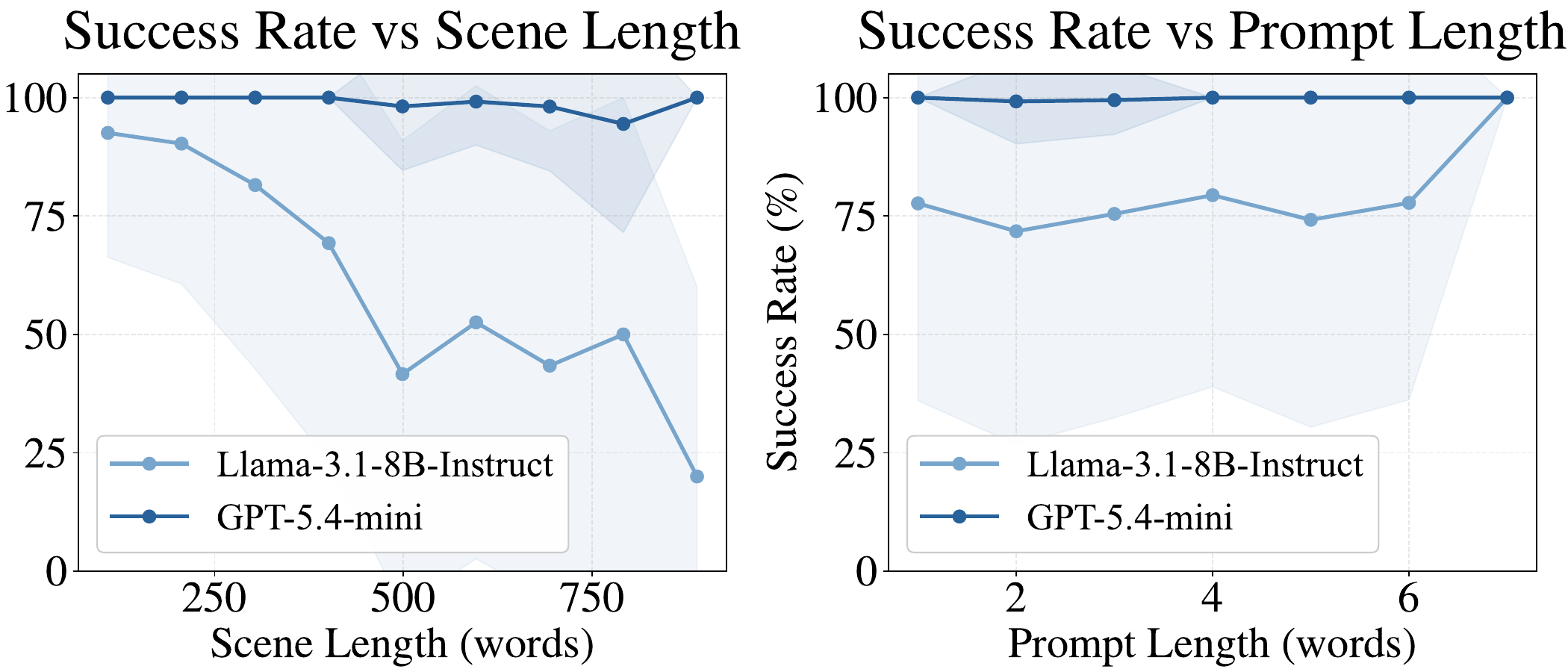}
    \vspace{-5mm}
    \caption{Removal accuracy on the \texttt{`liv'} dataset: GPT-5.4-mini (dark blue) vs.\ Llama-3.1-8B (light blue) across SSR length and prompt length.}
    \label{fig:removal-analysis-supp}
\end{figure}

\textbf{SSR Length Impact.} Figure \ref{fig:removal-analysis-supp} (left) shows removal accuracy as a function of SSR word count. We observe a dramatic drop in performance: scenes with <200 words achieve 95\% accuracy, while scenes with >500 words drop below 35\%. This strongly confirms that longer token sequences present a clear challenge for the 8B instruction-tuned model (Llama-3.1-8B-Instruct) used for removal. The \texttt{`liv'} split contains significantly more objects per scene, resulting in longer SSRs and explaining the performance gap. We hypothesize that larger and better instruction-tuned models would better handle long-context JSON manipulation. We confirm this directly: replacing Llama-3.1-8B with a stronger frontier model (GPT-5.4-mini) raises removal accuracy to $99.8\% \pm 0.2$ with no systematic dependence on SSR length and no recurring failure pattern, establishing that the degradation stems from the 8B model's long-context limits rather than any framework constraint.


\textbf{Prompt Length and Ambiguity.} Figure \ref{fig:removal-analysis-supp} (right) examines accuracy versus object prompt length. Longer prompts (7 words) achieve 100\% accuracy compared to $\sim$75\% for shorter prompts, suggesting that more specific descriptions reduce ambiguity. However, accuracy remains relatively flat from 1-6 words ($\sim$75\%), indicating that prompt ambiguity can not be the only reason.

\textbf{Failure Mode Analysis.} Categorizing the 326 total failures on the \texttt{`liv'} test set reveals that the primary bottleneck is \textit{different class} errors (182 failures, 56\%), where the model removes an object of the wrong class entirely (e.g., removing a table when prompted to remove a chair), a reasoning failure rather than semantic ambiguity. A further 115 \textit{same class} failures (35\%) occur when multiple objects of the requested class exist but the wrong instance is removed, attributable to both prompt ambiguity and our evaluation criterion, which requires all objects sharing the ground-truth \texttt{`desc'} to be removed for correctness. The remaining 29 failures (9\%) remove both correct and incorrect objects simultaneously. These wrong-class errors dominate for the 8B backbone but are almost entirely absent under the frontier model, confirming they stem from limited long-context reasoning rather than the task or representation.

These results indicate that removal accuracy is primarily limited by the long-context reasoning capacity of the 8B model rather than prompt ambiguity. The frontier-model result confirms this: the wrong-class errors that dominate Llama-3.1-8B failures resolve almost entirely with a stronger backbone, establishing that they reflect backbone capacity rather than a limitation of the task or our representation --- consistent with removal being an identification-and-deletion operation that scales with the zero-shot LLM.

\subsection{Example of SSR Instance}
\label{supp:ssr-example}
We show a full example of a Structured Scene Representation (SSR) instance with sampled assets in Listing \ref{ssr-instance}. The ``abstract” SSR—before concrete 3D asset selection—would simply not contain the key/value pairs with \texttt{`sampled\_'} prefix and the optional \texttt{`uuid'} key/value pair, as they are added after asset selection. For numerical values, we omit `pretty formatting' (with line breaks after every element) in order to fit the example into a single page in this PDF.

\begin{lstlisting}[caption={Example of SSR instance with sampled assets},label={ssr-instance}, language=json,firstnumber=1]
{
  "room_type": "bedroom",
  "bounds_top": [[-1.55, 2.6, 1.9], [1.55, 2.6, 1.9], [1.55, 2.6, -1.9], [-1.55, 2.6, -1.9]],
  "bounds_bottom": [[-1.55, 0.0, 1.9], [1.55, 0.0, 1.9], [1.55, 0.0, -1.9], [-1.55, 0.0, -1.9]],
  "objects": [
    {
      "desc": "A contemporary king-size bed with a brown padded headboard, Hello Kitty-themed pink and white bedding, graphic pillows, and bolster cushions, offering a comfortable aesthetic",
      "size": [ 1.77, 0.99, 1.94 ],
      "pos": [ 0.44, 0.0, -0.44 ],
      "rot": [ 0.0, 0.70711, 0.0, -0.70711 ],
      "jid": "8a31d51c-2306-439f-90c6-650be7284975",
      "sampled_asset_jid": "7bf721bf-8839-4343-95c5-b6e852805ad1",
      "sampled_asset_desc": "Modern minimalist king-size bed with a wood frame, padded gray fabric headboard, and clean lines.",
      "sampled_asset_size": [1.77, 1.02, 2.03],
      "uuid": "d3d31dbc-ff1d-4122-8a80-52598c326f00"
    }, ... ]
}
\end{lstlisting}

\subsection{Prompts for Zero-Shot Model}
\label{supp:prompts-zero-shot}
The full prompt for user instruction decomposition is in Figure \ref{fig:respace-main-prompt}. We further show an example of input/output prompts for a full scene generation in Figure \ref{fig:prompt-example-full-scene}. The prompt for object removal, using the same zero-shot LLM, is shown in Figure \ref{fig:respace-removal-prompt}, with an example of input/output in Figure \ref{fig:prompt-example-removal-scene}.

\begin{figure*}
    \begin{tcolorbox}[top=2pt,bottom=2pt, width=\linewidth, boxrule=1pt, halign=left]
        {\scriptsize \setstretch{0.9} {\fontfamily{zi4}\selectfont
        \textbf{System Prompt} \\
you are a world-class leading interior design expert. your task is to fulfill the request of the user about interior design but you have help of another world-class expert model that can only be called in an XML-style API.
 
\# input

- <prompt> : the user request

- <scenegraph> : the current scene will be given as a JSON object. in some cases, there will be no scene graph given, which means there is no ``current" scene to work with. the ``bounds\_top" and ``bounds\_bottom" keys contain the boundaries as a list of 3D vertices in metric space.

\# task

- composing a list of commands to fulfill the user request via <add> and <remove> commands. ideally, you reflect the existing objects in the scenegraph, if one is given.

\# adding

- if the user wants to add one or multiple objects, you create an <add> command for every object/furniture and add it to the list in ``commands". 

- for the description, you should refer to the subject with a maximum of five additional descriptive words. the first words should refer to the color / style / shape / etc., while the last word should always be the main subject. your description must be in `noun phrase'.

- if the user request provides an existing scene description provided via <scenegraph>...</scenegraph> and there are existing objects in the scene, you should try to match the style of the existing objects by providing a similar style as part of the description of your commands.

- if the user provides some requirement about particular furniture that should be present in the room, you should always add these objects via <add> commands.

- your format should be: <add>description</add>

- DO NEVER use more than 5 words for each description

\# removing / swapping

- if the user wants to remove one to multiple objects, you add a <remove> command for every object that should be removed.

- if the user wants to swap or replace furniture, you MUST use <remove> first and then use <add>

- if there are similar candidates for removal you should remove the object that matches the description best.

- your format should be: <remove>description</remove>

- you can keep the description short here as well

\# output

- the commands are given as a list under the ``commands" key where each command follows EXACTLY the format specified above and is given as a string, i.e. ``<add>...</add>" or ``<remove>...</remove>".

- if there are remove commands, you always put them BEFORE add commands.

- IMPORTANT: you NEVER use the <remove> commands unless the user EXPLICITLY asks for it via swapping or removing objects. you do not make assumptions about this.

- you NEVER remove objects to "match the style" or if there is already an object in the scene similar to the requested one. a scene can contain as many similar objects as the user wants. you ONLY remove objects if the user explicitly asks for removal or swapping.

- if you use the <remove> command, you MUST provide your reasoning under the "reasoning" key, which comes before the "commands" key in the same JSON object.
- you always output the final JSON object as a plain string and nothing else. NEVER use markdown.

\# available object classes

- you should only pick objects for <add> based on the following high-level abstract classes

- your objects should be more specific than these classes but you should not add objects that are not part of these classes/labels

\{UNIQUE\_OBJECT\_CLASSES\}

\# available object classes

- you should only pick objects for <add> based on the following high-level abstract classes

- your objects should be more specific than these classes but you should not add objects that are not part of these classes/labels

\# few-shot examples for scenes that have a similar size to the requested one (your scene should be different though and stick to the user prompt):\{PROMPT\_LISTS\_FOR\_K\_EXAMPLES\}

REMINDER: each description in your <add>...</add> commands should be IN NOUN PHRASE WITH 2-3 words AND AT MAXIMUM 5 words
        }
        \par}
    \end{tcolorbox}
    \begin{tcolorbox}[top=2pt,bottom=2pt, width=\linewidth, boxrule=1pt, halign=left]
        {\scriptsize \setstretch{1.0} {\fontfamily{zi4}\selectfont
        \textbf{User Prompt} \\
        <prompt>\{UNSTRUCTURED\_USER\_INSTRUCTION\}<prompt> \\
        <scenegraph>\{JSON\_DUMP\_OF\_SSR\_IF\_PROVIDED\_OR\_NONE\}</scenegraph>
        }
        \par}
    \end{tcolorbox}
    \vspace{-2mm}
    \caption{System and User Prompt for the zero-shot LLM for command decomposition.}
    \label{fig:respace-main-prompt}
\end{figure*}

\begin{figure*}
    \begin{tcolorbox}[top=2pt,bottom=2pt, width=\linewidth, boxrule=1pt, halign=left]
        {\scriptsize \setstretch{1.0} {\fontfamily{zi4}\selectfont
        \textbf{User Prompt} \\
        <prompt>create a bedroom with 5 objects.<prompt>
<scenegraph>{``room\_type": ``bedroom", ``bounds\_top": [[-3.2, 2.65, 1.7], [3.2, 2.65, 1.7], [3.2, 2.65, -0.1], [0.9, 2.65, -0.1], [0.9, 2.65, -1.7], [-3.2, 2.65, -1.7]], ``bounds\_bottom": [[-3.2, 0.0, 1.7], [3.2, 0.0, 1.7], [3.2, 0.0, -0.1], [0.9, 0.0, -0.1], [0.9, 0.0, -1.7], [-3.2, 0.0, -1.7]], ``objects": []}</scenegraph>
        }
        \par}
    \end{tcolorbox}
    \begin{tcolorbox}[top=2pt,bottom=2pt, width=\linewidth, boxrule=1pt, halign=left]
        {\scriptsize \setstretch{1.0} {\fontfamily{zi4}\selectfont
        \textbf{Response (Model Output)} \\
        {`commands': [`<add>dark wooden double bed</add>', `<add>creamy white nightstand</add>', `<add>black floor lamp</add>', `<add>white two-seat sofa</add>', `<add>low shelf</add>']}
        }
        \par}
    \end{tcolorbox}
    \vspace{-2mm}
    \caption{Example of an input/output pair to the zero-shot LLM on full scene synthesis. Each command gets iteratively processed by ReSpace. For full scene synthesis, this results in an autoregressive loop into SG-LLM such that objects get added into the partial scene.}
    \label{fig:prompt-example-full-scene}
\end{figure*}

\begin{figure*}
    \begin{tcolorbox}[top=2pt,bottom=2pt, width=\linewidth, boxrule=1pt, halign=left]
        {\scriptsize \setstretch{1.0} {\fontfamily{zi4}\selectfont
        \textbf{System Prompt} \\
        you are a world-class leading interior design expert. your task is to remove furniture given the descriptions in the header and the current list of furniture in the body. you must respond ONLY with a valid JSON string that matches precisely the *format* of the existing JSON in the request.

if there are multiple objects that match the description precisely, you should remove all of them.

the prompt for the object to be removed will be given in the header between <remove>...</remove> tags. the current scene will be given as a JSON object in the body between <scenegraph>...</scenegraph> tags.

in the successful case, your output contains one or N fewer objects in the "objects" list and the rest of the JSON object should be EXACTLY identical to the input.

you can also remove all objects if the prompt matches those objects. in that case, you provide an empty list for the ``objects" key.

you can further assume that in most cases, there will be at least one object in the scene that matches the description roughly. this object shall be removed.

only output the JSON (with the removed objects) as a plain string and nothing else.
        }
        \par}
    \end{tcolorbox}
    \begin{tcolorbox}[top=2pt,bottom=2pt, width=\linewidth, boxrule=1pt, halign=left]
        {\scriptsize \setstretch{1.0} {\fontfamily{zi4}\selectfont
        \textbf{User Prompt} \\
        <remove>\{OBJECT\_PROMPT\_FOR\_REMOVAL\}<remove> \\
        <scenegraph>\{JSON\_DUMP\_OF\_SSR\}</scenegraph>
        }
        \par}
    \end{tcolorbox}
    \vspace{-2mm}
    \caption{System and User Prompt for the zero-shot LLM for object removal.}
    \label{fig:respace-removal-prompt}
\end{figure*}

\begin{figure*}
    \begin{tcolorbox}[top=2pt,bottom=2pt, width=\linewidth, boxrule=1pt, halign=left]
        {\scriptsize \setstretch{1.0} {\fontfamily{zi4}\selectfont
        \textbf{User Prompt} \\
        <remove>comfortable aesthetic bed<remove>\\
        <scenegraph>{``room\_type": ``bedroom", ``bounds\_top": [[-1.55, 2.6, 1.9], [1.55, 2.6, 1.9], [1.55, 2.6, -1.9], [-1.55, 2.6, -1.9]], ``bounds\_bottom": [[-1.55, 0.0, 1.9], [1.55, 0.0, 1.9], [1.55, 0.0, -1.9], [-1.55, 0.0, -1.9]], ``objects": [{``desc": ``Mid-Century Modern nightstand with light wood finish, geometric cutout handle, and angled legs.", ``size": [0.5, 0.55, 0.43], ``pos": [1.37, 0.0, -1.45], ``rot": [0, -0.70711, 0, 0.70711], ``jid": ``9603344b-99b8-43db-abf0-73c7eaf0ea5f-(0.81)-(1.0)-(0.72)"}, {``desc": ``Modern minimalist TV stand with walnut brown wood, white accents, two closed cabinets, and open shelving on raised legs.", ``size": [1.86, 0.42, 0.35], ``pos": [-1.35, 0.0, -0.52], ``rot": [0, 0.70711, 0, 0.70711], ``jid": "18d54650-68ae-4d4b-8079-1f050b267153-(1.01)-(1.0)-(0.82)"}, {``desc": "Modern industrial wardrobe with a minimalist metal and mesh design, featuring a rectangular shape, four slender legs, a single shelf, and a hanging rod.", ``size": [1.6, 2.11, 0.5], ``pos": [0.74, 0.0, 1.56], "rot": [0, 1, 0, 0], ``jid": ``19035101-21a1-4495-ae95-90d8d1ccd108-(1.55)-(1.0)-(1.04)"}, {``desc": ``Modern minimalist floor lamp with a white fabric drum shade and brown wooden tripod base featuring an open geometric design.", ``size": [0.47, 1.1, 0.47], ``pos": [-1.27, 0.0, 0.6], ``rot": [0, -0.35334, 0, 0.93549], ``jid": "4e086af8-0900-4c6d-b535-08c7a936a981-(1.0)-(0.85)-(1.0)"}, {``desc": "Modern adjustable pendant lamp with branching black metal arms, gold disk shades, and minimalist design.", ``size": [0.36, 1.03, 0.32], ``pos": [0.06, 1.79, -0.13], ``rot": [0, 0, 0, 1], ``jid": "6289a052-b4a7-417a-b687-b22fddc14663-(0.23)-(1.0)-(0.31)"}, {``desc": "Mid-Century Modern nightstand with light wood finish, geometric cutout handle, and angled legs.", ``size": [0.5, 0.55, 0.43], ``pos": [1.35, 0.0, 0.62], ``rot": [0, -0.70711, 0, 0.70711], ``jid": "9603344b-99b8-43db-abf0-73c7eaf0ea5f-(0.81)-(1.0)-(0.72)"}, {``desc": ``A contemporary king-size bed with a brown padded headboard, Hello Kitty-themed pink and white bedding, graphic pillows, and bolster cushions, offering a playful yet comfortable aesthetic.", ``size": [1.56, 0.9, 2.22], ``pos": [0.54, 0.0, -0.43], ``rot": [0, -0.70711, 0, 0.70711], ``jid": ``bff6e7b4-4e17-401d-b684-dc768def3869-(0.88)-(0.91)-(1.14)"}]}</scenegraph>
        }
        \par}
    \end{tcolorbox}
    \begin{tcolorbox}[top=2pt,bottom=2pt, width=\linewidth, boxrule=1pt, halign=left]
        {\scriptsize \setstretch{1.0} {\fontfamily{zi4}\selectfont
        \textbf{Response (Model Output)} \\
        {``room\_type": ``bedroom", ``bounds\_top": [[-1.55, 2.6, 1.9], [1.55, 2.6, 1.9], [1.55, 2.6, -1.9], [-1.55, 2.6, -1.9]], ``bounds\_bottom": [[-1.55, 0.0, 1.9], [1.55, 0.0, 1.9], [1.55, 0.0, -1.9], [-1.55, 0.0, -1.9]], ``objects": [{``desc": ``Mid-Century Modern nightstand with light wood finish, geometric cutout handle, and angled legs.", ``size": [0.5, 0.55, 0.43], ``pos": [1.37, 0.0, -1.45], ``rot": [0, -0.70711, 0, 0.70711], ``jid": ``9603344b-99b8-43db-abf0-73c7eaf0ea5f-(0.81)-(1.0)-(0.72)"}, {``desc": ``Modern minimalist TV stand with walnut brown wood, white accents, two closed cabinets, and open shelving on raised legs.", ``size": [1.86, 0.42, 0.35], ``pos": [-1.35, 0.0, -0.52], ``rot": [0, 0.70711, 0, 0.70711], ``jid": "18d54650-68ae-4d4b-8079-1f050b267153-(1.01)-(1.0)-(0.82)"}, {``desc": "Modern industrial wardrobe with a minimalist metal and mesh design, featuring a rectangular shape, four slender legs, a single shelf, and a hanging rod.", ``size": [1.6, 2.11, 0.5], ``pos": [0.74, 0.0, 1.56], "rot": [0, 1, 0, 0], ``jid": ``19035101-21a1-4495-ae95-90d8d1ccd108-(1.55)-(1.0)-(1.04)"}, {``desc": ``Modern minimalist floor lamp with a white fabric drum shade and brown wooden tripod base featuring an open geometric design.", ``size": [0.47, 1.1, 0.47], ``pos": [-1.27, 0.0, 0.6], ``rot": [0, -0.35334, 0, 0.93549], ``jid": "4e086af8-0900-4c6d-b535-08c7a936a981-(1.0)-(0.85)-(1.0)"}, {``desc": "Modern adjustable pendant lamp with branching black metal arms, gold disk shades, and minimalist design.", ``size": [0.36, 1.03, 0.32], ``pos": [0.06, 1.79, -0.13], ``rot": [0, 0, 0, 1], ``jid": "6289a052-b4a7-417a-b687-b22fddc14663-(0.23)-(1.0)-(0.31)"}, {``desc": "Mid-Century Modern nightstand with light wood finish, geometric cutout handle, and angled legs.", ``size": [0.5, 0.55, 0.43], ``pos": [1.35, 0.0, 0.62], ``rot": [0, -0.70711, 0, 0.70711], ``jid": "9603344b-99b8-43db-abf0-73c7eaf0ea5f-(0.81)-(1.0)-(0.72)"}]}
        }
        \par}
    \end{tcolorbox}
    \vspace{-2mm}
    \caption{Example of an input/output pair to the zero-shot LLM on object removal.}
    \label{fig:prompt-example-removal-scene}
\end{figure*}

\end{document}